\documentclass[lettersize,journal]{IEEEtran}
\usepackage{amsmath,amsfonts}
\usepackage{algorithmic}
\usepackage{algorithm}
\usepackage{array}
\usepackage[caption=false,font=normalsize,labelfont=sf,textfont=sf]{subfig}
\usepackage{textcomp}
\usepackage{stfloats}
\usepackage{url}
\usepackage{verbatim}
\usepackage{graphicx}
\usepackage{cite}
\usepackage{amsmath,amssymb,amsfonts,dsfont,pifont,bm,bbm,mathrsfs,mathtools,nicefrac, makecell, multirow}
\usepackage{float}
\usepackage{overpic}
\usepackage{marvosym}
\usepackage{hyperref}
\usepackage{cleveref}
\crefname{figure}{Fig.}{figures}
\crefname{section}{Sec.}{sections}
\crefname{equation}{Eq.}{Eqs.}
\crefname{table}{Tab.}{Tabs.}
\usepackage{booktabs}
\begin{document}

\title{OmniPrism: Learning Disentangled Visual Concept for Image Generation}

\author{Yangyang Li, Wu Liu,~\IEEEmembership{Senior Member,~IEEE,} Daqing Liu, Allen He, Xinchen Liu,~\IEEEmembership{Member,~IEEE,} Guoqing Jin, \\ Yongdong Zhang,~\IEEEmembership{Fellow, IEEE} 
        % <-this % stops a space
\thanks{Yangyang Li, Wu Liu, Yongdong Zhang are with the School of Cyber Science and Technology, at the University of Science and Technology of China, Anhui, China. (email: lyy1030@mail.ustc.edu.cn, \{liuwu, zyd73\}@ustc.edu.cn).}% <-this % stops a space
\thanks{Daqing Liu, Xinchen Liu, Allen He are with the JD Explore Academy, at JD.com Inc, Beijing, China. (email: \{daqing.liu, allenhethis\}@outlook.com, liuxinchen1@jd.com).}% <-this % stops a space
\thanks{Guoqing Jin is with the State Key Laboratory of Communication Content Cognition, at People’s Daily Online, Beijing, China. (email: jinguoqing@people.cn.}% <-this % stops a space
% \thanks{Manuscript received April 19, 2021; revised August 16, 2021.}
}
% The paper headers
% \markboth{Journal of \LaTeX\ Class Files,~Vol.~14, No.~8, August~2021}%
% {Shell \MakeLowercase{\textit{et al.}}: A Sample Article Using IEEEtran.cls for IEEE Journals}

% \IEEEpubid{0000--0000/00\$00.00~\copyright~2021 IEEE}
% % Remember, if you use this you must call \IEEEpubidadjcol in the second
% % column for its text to clear the IEEEpubid mark.

\maketitle

\begin{abstract}
Creative visual concept generation often draws inspiration from specific concepts in a reference image to produce relevant outcomes.
However, existing methods are either constrained to single-dimensional concept generation and difficult to simply apply to concepts of other dimensions, or are easily disturbed by irrelevant concepts in multi-dimensional concept generation, resulting in concept leakage.
To address these problems, we propose \textbf{OmniPrism}
%\footnote{\textbf{OmniPrism} is inspired by the way a prism disperses light into its component, analogous to how we disentangle concepts from an image.}, 
%
a visual concept disentangling approach for disentangled multi-dimension visual concept generation.
We utilize the rich semantic space of a multimodal extractor to disentangle images into orthogonal concept dimension (e.g., content, style, layout) driven by language and learning these concept representations via contrastive orthogonal disentangled (COD) in diffusion models. 
To focus on the target concept during training, we construct the first paired concept disentangled dataset (PCD-200K), where each pair shares the same target concept and other irrelevant concepts.
And a set of block embeddings is designed to adapt each diffusion block's concept domain to achieve better concept consistency.
Extensive experiments demonstrate that our method can generate high-quality, concept-disentangled results with high fidelity to text prompts and desired concepts, promoted the development of creative visual concept generation.
\end{abstract}

    
\section{Introduction}
\label{sec:intro}

\IEEEPARstart{V}{isual} concept generation aims to generate specific visual elements from images, which demonstrates significant value in artistic creation and graphic design. Recent research about text-to-image (T2I) generation based on diffusion models~\cite{nichol2021glide, ramesh2022hierarchical, rombach2022high, saharia2022photorealistic} achieve remarkable progress in visual concept generation, providing controllability for generating the concepts that people desired from reference images. The visual concepts in images can be disentangled into three orthogonal dimensions: content (semantic subjects), style (artistic styles, colors, \textit{etc.}), and layout (global relationships, pose of local subject, \textit{etc.}). The independent or combined control of these concepts provide broad applications. However, how to flexibly disentangle and generate different concepts remains a challenge.

\begin{figure}[t]
    \centering
    \includegraphics[width=0.48\textwidth]{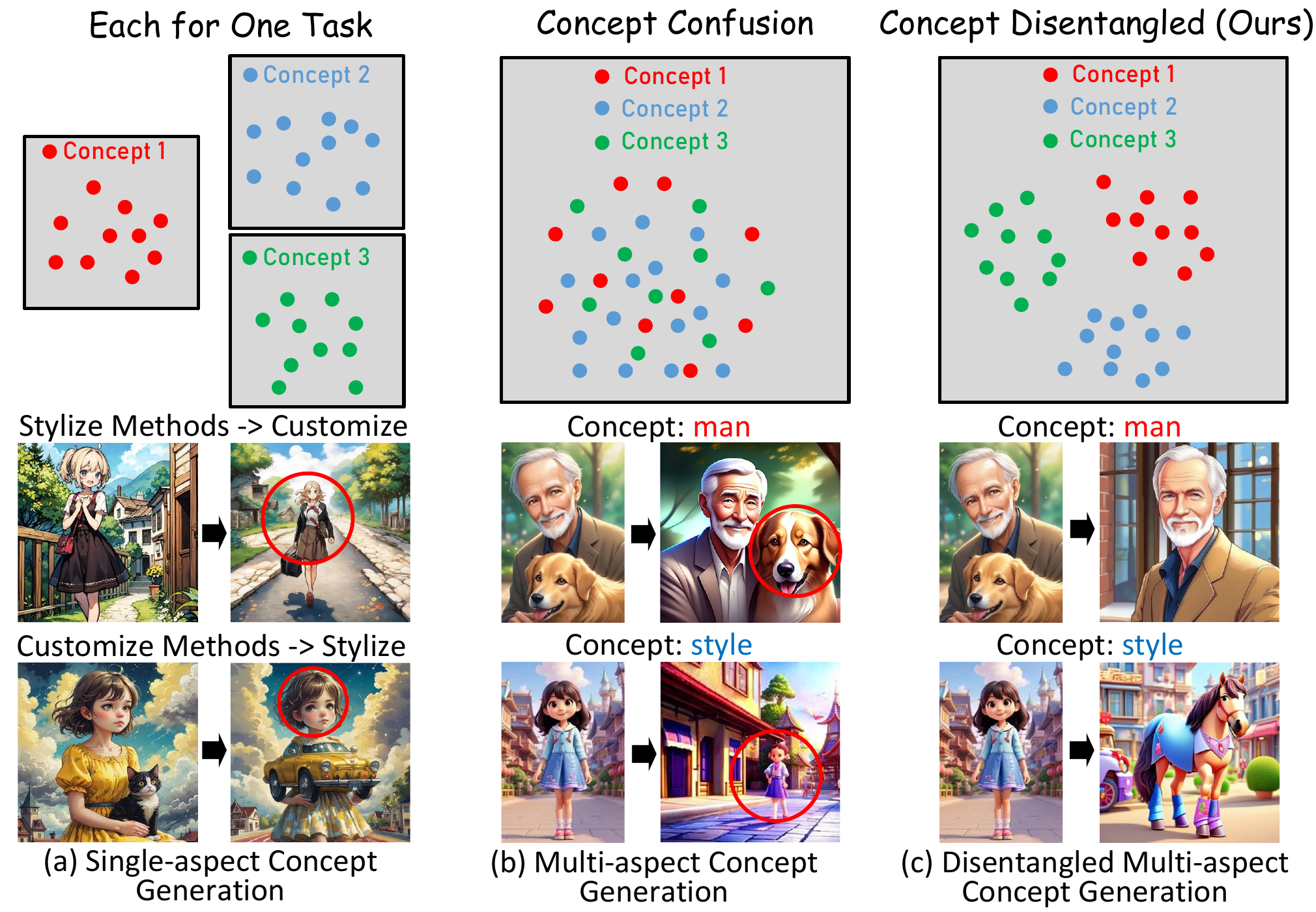}
    \caption{%
    \textbf{Challenges in visual concept generation.} (a) Limited concept space of single-dimension concept generation, which is only suitable for single tasks. (b) Previous multi-dimension concept generation works often struggled with concept leakage. (c) We disentangle different concepts in the representation space, thereby achieving results without irrelevant concepts.
    }
    \label{fig:problem}
    \vspace{-10pt}
\end{figure}

Existing research mostly focus on the single-dimension concept generation, such as customization~\cite{gal2023an, ruiz2023dreambooth, wei2023elite, kumari2022customdiffusion, saharia2022photorealistic, chen2024anydoor, shi2024instantbooth, zhang2024two, chen2025videodreamer,jiang2024animediff}, stylization~\cite{sohn2024styledrop, wang2023styleadapter, xu2024sgdm,cao2023difffashion, hertz2024style, qi2024deadiff, jeong2024visual}, or structural reference generation~\cite{huang2023reversion, huang2024learning, zhang2023adding}. However, such methods are limited in: 1) they are usually designed for specific concept dimension (such as the subject or style), and it is difficult to generate the expected results when simply applied to other concept dimensions, such as stylization model cannot generate the target girl in~\cref{fig:problem} (a); 2) some works~\cite{avrahami2023break, chen2023disenbooth} rely on complex preprocessing (\textit{e.g.} subject masks) or online fine-tuning, and it is difficult to generalize them to abstract concepts (\textit{e.g.} style or relation between subjects), or to scenarios that require fast generation. These limitations stem from the neglect of the entangling characteristics among concepts in single-dimension modeling, resulting in a lack of cross-concept controllability in the generated results.

To break through these limitations, some recent works attempt to generate multi-dimension concepts under language-driven guidance. Xu~\textit{et al.}~\cite{xu2024cusconcept} and Yael~\textit{et al.}~\cite{vinker2023concept} bind multi-dimension visual concepts with discrete language tokens to establish cross-modal association. However, they have difficulty in capturing the fine-grained concept representations, resulting in low concept consistency. Other works introduce multimodal encoder to align visual concept representations to the language space such as BLIP-Diffusion~\cite{li2024blip}, DEADiff~\cite{qi2024deadiff}, SSR-Encoder~\cite{zhang2024ssr} and MS-Diffusion~\cite{wang2025msdiffusion}, they then learn these representations through cross-attention. OmniGen~\cite{xiao2024omnigen} further advances this paradigm by implementing a multimodal Diffusion Transformer (DiT)~\cite{peebles2023scalable} framework to achieve unified concept generation. Nevertheless, these methods lack explicit constraints on concepts in different dimensions within the concept representation space, causing concept entanglement and are easily interfered by needless concepts during generation, such as the concept leakage shown in~\cref{fig:problem} (b).

\begin{figure*}[ht]
    \centering
    \includegraphics[width=\linewidth]{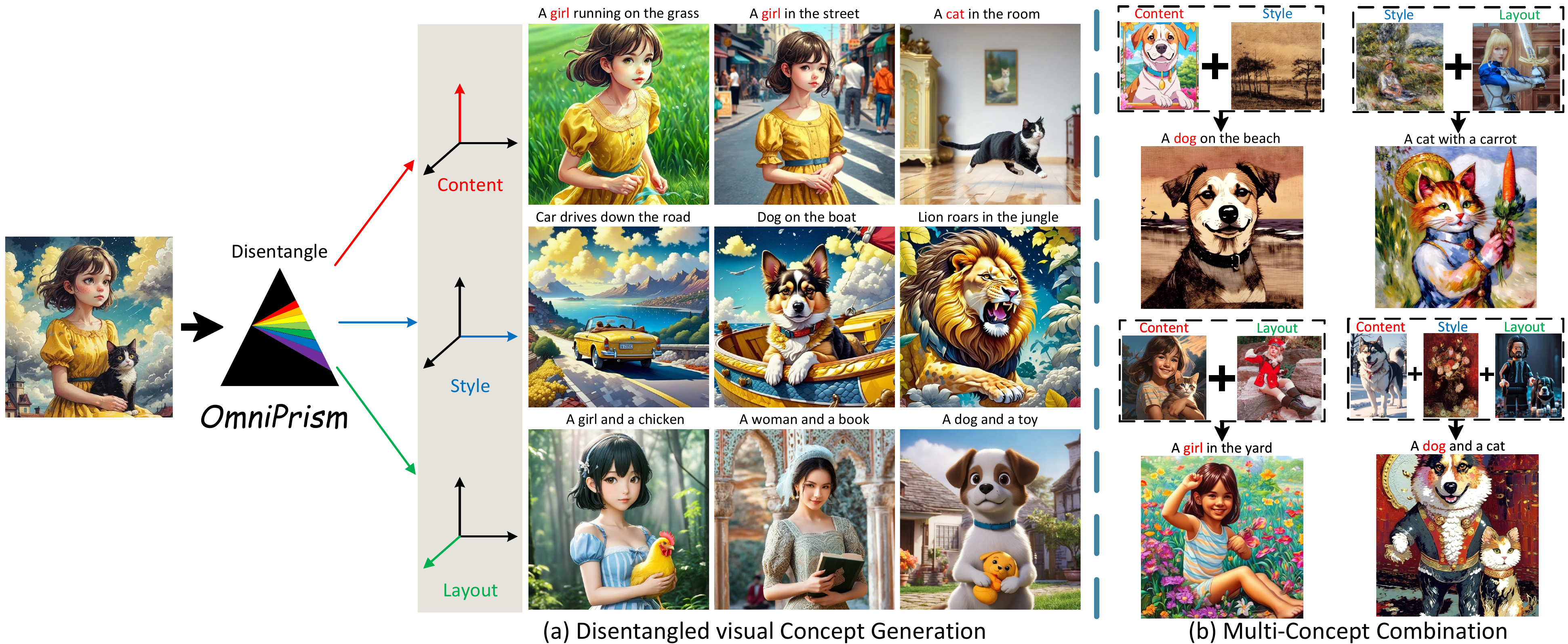}
    \caption{%
        We propose \textbf{OmniPrism}, which arbitrarily disentangles and combines visual concepts. (a) Disentangled visual concept generation. 
        Given a reference image with multiple concepts, our method can disentangle the desired concept guided by natural language such as content names (red color words in prompts), ``style'' or ``layout" (\textit{e.g.}, global relationship or pose of local subject) while remaining faithful to prompts. (b) Multi-concept combination. Given two or more reference images with the corresponding concept guidance, our approach can combine all desired concepts in any combination without conflicts.
        }
    \label{fig:fig1}
    % \vspace{-10pt}
\end{figure*}

The above problems are due to two core challenges. Firstly, the prominence of target concepts in the training data is insufficient, which leads the model to tend to reconstruct the reference images rather than specifically extracting the target concept. Secondly, different concepts lack effective distinguishability in the representation space and are easily interfered by irrelevant concepts during generation. To address these challenges, we propose a data-driven orthogonal disentangling framework based on contrastive learning. By explicitly constructing positive and negative sample pairs, the perception of target concepts is strengthened. Meanwhile, we achieve the free disentangled generation of multi-dimensional concepts by contrastive learning mechanism.

Specifically, we propose a visual concept generation approach based on Contrastive Orthogonal Disentangled (COD) learning, named \textbf{OmniPrism}, as shown in~\cref{fig:fig1}. Firstly, we adopt a multimodal representation extractor based on the learnable Q-Former~\cite{li2023blip} for its rich semantic space. Secondly, we design the COD learning mechanism. By constraining the target concepts to be close to similar concepts and orthogonal to irrelevant concepts in the representation space, we achieve fine-grained concept disentanglement. Finally, we introduce a set of block embeddings corresponding to every diffusion blocks. These embeddings are respectively added to the concept extraction process and obtain representations aligned with each block's concept domain, thus improving the generated concept consistency with target concepts.
To support the model training, we construct the first Paired Concept Disentanglement dataset (PCD-200K) containing 200,000 paired samples. Each pair contains a shared concept and other distinct concepts, which provides a crucial data basis for concept disentangled learning.
Experiments show that we can accurately disentangle diverse concepts. Moreover, benefiting from the orthogonality of concepts representations, we can directly achieve the free combination of heterogeneous concepts across images, providing a new path for the controllable combination of multi-dimensional visual concepts, as shown in~\cref{fig:fig1} (b).

Our main contributions are summarized as follows:
\begin{itemize}
\item We propose \textbf{OmniPrism}, a novel disentangled visual concept generation approach with high fidelity to text prompts and desired concepts.
\item We propose a Contrastive Orthogonal Disentangle (COD) Learning mechanism to disentangle concepts guided by natural language, and a novel block embedding to adapt the concept domain of each diffusion block.
\item We construct the Paired Concept Disentanglement dataset (PCD-200K) for concept disentanglement, where each sample contains the same concept and irrelevant concepts.
\end{itemize}

\section{Related Work}
\label{sec:related}

\subsection{Text-to-Image Diffusion Models}
Recently, diffusion models have achieved great success in text-to-image generation.
Based on classifier-free guidance \cite{ho2022classifier}, GLIDE \cite{nichol2021glide}, DALLE2 \cite{ramesh2022hierarchical}, Stable Diffusion \cite{rombach2022high} and Imagen \cite{saharia2022photorealistic} introduced large-scale multi-modal models such as CLIP~\cite{radford2021learning}, and achieved significant improvements in text-conditional image generation of open vocabulary. Some improvements have been made to improve the quality of T2I generation in all aspects.
Stable Diffusion XL \cite{podell2023sdxl} achieves high-resolution results on larger models and resolves problems with aspect ratio and cropping in generated images. Recently, Stable Diffusion 3 \cite{esser2024scaling} and Flux \footnote{https://github.com/black-forest-labs/flux} based on the diffusion transformer (DiT) \cite{peebles2023scalable} achieve higher-quality image generation.
These models are designed only for generation controlled by text condition. However, images contain richer semantic information than text, and image-conditioned models have more controllable generation capabilities. Therefore, how to reference images for generation is a meaningful task.

\subsection{Single-Dimension Visual Concept Generation}
Visual concept generation based on diffusion models is typically conditioned on both image and text, aiming to generate results that incorporate visual concepts from image (subject, style, relation, \textit{etc.}) while faithful to text prompts.
Text Inversion~\cite{gal2022image} optimizes an extra embedding for a reference image in the text domain to generate similar subjects. 
DreamBooth~\cite{ruiz2023dreambooth} binds a unique identifier in the token domain to the reference image by optimizing the model and generate customized results. 
ControlNet~\cite{zhang2023adding} adds an additional control branch to learn the structural representations (\textit{e.g.} edges, depth) from the reference images and generate corresponding results. Some encoder-based methods~\cite{yang2023paint, wei2023elite, chen2023anydoor} introduce image encoders (VIT~\cite{dosovitskiy2021an}, DINO~\cite{zhang2023dino}, etc.) to extract representations of reference images and train a model to fit them for training-free generation.
IP-Adapter~\cite{ye2023ip} designs a set of plug-and-play adapters to help the models learn image representations without compromising its original capabilities. 
Other works injects different conditions into different U-Net blocks~\cite{voynov2024p,agarwal2023image,qi2024deadiff,wang2024instantstyle} or different time steps~\cite{zhang2023prospect,agarwal2023image} during inference to separate concepts.
However, these methods are limited to single concept and cannot to be flexibly applied to multiple concepts, leading to concept conflict.
Lego~\cite{motamed2024lego} lets the target content away from its antonym in text space to disentangle it, but the lack of alignment in image space limits their identity preservation.
In contrast, our method uses language as guidance to disentangle desired concepts in image space, thereby achieving accurate multi-dimension visual concept generation without concept conflict.

\subsection{Multi-Dimension Visual Concept Generation}
Compared with single-dimension visual concept generation, multi-dimension visual concept generation incorporates additional conditions as guidance to flexibly generate the desired concepts in the reference images.
Liu et al.~\cite{liu2014vehicle} learned consistent visual representations through global graph modeling and multimodal data fusion, achieving accurate modeling of target identity and structural relationships.
Some inference-tuning works~\cite{avrahami2023break, zhang2024attention, jin2024an} using additional spatial control (mask, layout, \textit{etc.}) or multiple reference images, binding the desired concept to a language mnemonic to achieve concept disentanglement. However, these methods are time-consuming for learning each sample and require complex control conditions, which limits their practical application.
Xu~\textit{et al.}~\cite{xu2024cusconcept} and Yael~\textit{et al.}~\cite{vinker2023concept} bind the concepts and attributes of an image to a set of tokens, but their concept similarity in generated results is limited.
HumanNeRF-SE~\cite{ma2024humannerf} utilizes SMPL priors to achieve pose-appearance disentangling, but is limited to human-centric scenarios.
MAPLE~\cite{chen2022maple} combines mask autoencoders with pseudo-labels to learn decoupled visual representations. However, it focuses on perceptual tasks and cannot introduce decoupled representations into image generation.

Other works use a multimodal encoder to extract concept representations.
Blip-Diffusion~\cite{li2024blip} employs Q-Former~\cite{li2023blip} to jointly learn image features and text features of subject name. However, they only use this multi-modal interaction to enhance visual representations and cannot disentangle visual concepts described by text.
DEADiff~\cite{qi2024deadiff} uses ``content'' and ``style'' as guidance to extract representations and inject them into certain U-Net layers to disentangle style concepts. However, they cannot disentangle concepts except style since their limited language domain. 
SSR-Encoder~\cite{zhang2024ssr} aligns image representations with language guidance in the CLIP~\cite{radford2021learning} representations space and generate corresponding concept as language guidance. 
MS-Diffusion~\cite{wang2025msdiffusion} introduces grounding resampler and multi-subject attention to enable multiple subjects combination in an image. OmniGen~\cite{xiao2024omnigen} designs a multimodal Diffusion Transformer (DiT) architecture~\cite{peebles2023scalable} to achieve unified concept generation.
But they all suffer from the trade-off of concept fidelity and concept independence, since their language space lacks the property of concept disentanglement.
Different from these works, our approach disentangles different concept representations by contrastive orthogonal disentangled learning, thereby generating desired concepts without concept leakage and achieving disentangled multi-dimension visual concept generation.    
\section{Method}
\label{sec:method}

\begin{figure*}[ht]
    \centering
    \includegraphics[width=0.95\textwidth]{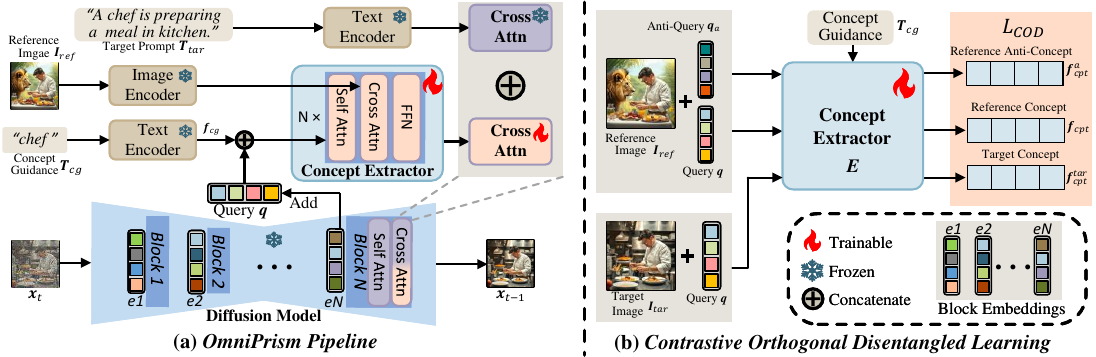}
    \caption{%
    \textbf{Framework of OmniPrism.} (a) Given the reference image $\bm{I}_{ref}$, target prompt $\bm{T}_{tar}$ and concept guidance $\bm{T}_{cg}$, the concept extractor disentangles concept representations $\bm{f}_{cpt}$ by concatenating CLIP features $\bm{f}_{cg}$ of $\bm{T}_{cg}$ with a learnable query $\bm{q}$, and feeds $\bm{f}_{cpt}$ into additional cross-attention layers in U-Net to generate target image $\bm{I}_{tar}$. A learnable block embedding $\bm{e}_i$ is added to $\bm{q}$ to align the concept domain of i-th diffusion block. (b) We employ an anti-query $\bm{q}_a$ to capture irrelevant concepts $\bm{f}^{a}_{cpt}$ in $\bm{I}_{ref}$, and constrain the desired concept $\bm{f}^{tar}_{cpt}$ in $\bm{I}_{tar}$ to be similar to $\bm{f}_{cpt}$ and orthogonal to $\bm{f}^{a}_{cpt}$ by Contrastive Orthogonal Disentangled (COD) Learning. 
    }
    \label{fig:framework}
\end{figure*}

Given a text prompt $\bm{T}_{tar}$, a reference image $\bm{I}_{ref}$, and a concept guidance $\bm{T}_{cg}$, our goal is to generate an image $\bm{I}_{tar}$ that is faithful to $\bm{T}_{tar}$ and incorporates the desired concept without irrelevant concepts in $\bm{I}_{ref}$. To achieve this, we first construct a Paired Concept-Disentangled (PCD-200K) dataset with all these data (\cref{sec:dataset}), then propose a disentangled visual concept generation approach termed \textbf{OmniPrism}, which employs a concept extractor $\bm{E}$ to disentangle specific concept representations $\bm{f}_{cpt}$ from $\bm{I}_{ref}$ guided by $\bm{T}_{cg}$ and feed $\bm{f}_{cpt}$ into additional cross-attention layers of diffusion models (\cref{sec:cdip}). A set of block embeddings $\bm{e}_1$ to $\bm{e}_N$ (\cref{sec:cdip}) are designed to align the concept domain of each diffusion blocks, $N$ is the number of diffusion blocks. We design a Contrastive Orthogonal Disentangled (COD) learning mechanism (\cref{sec:contrastive}) to train our model. 

\subsection{Preliminary}
\label{sec:pre}
Diffusion models~\cite{rombach2022high} are generative models that use a network $\boldsymbol\epsilon_\theta$ to gradually denoise random Gaussian noise $\bm{z}_T\sim \mathcal{N}(0,1)$ to learn the data distribution. This is the reverse process of adding noise to the image $\bm{z}_0$ by $T$ steps Markov chain. Diffusion models use text prompts $\bm{c}$ as conditions for text-to-image generation, and the training objective is to predict the noise $\boldsymbol\epsilon$ added to the noised latent $\bm{z}_t$ at time step $t$, which can be simplified as a variant of the variational bound:
\begin{equation}\label{eq:rec_loss}
\mathcal{L}_{ldm} = \mathbb{E}_{\bm{z}, \bm{c}, \boldsymbol\epsilon \sim \mathcal{N}(0,1), t}[\|\boldsymbol\epsilon_\theta(\bm{z}_t, \bm{c}, t)-\boldsymbol\epsilon\|_2^2].
\end{equation}
During inference, $\boldsymbol\epsilon_\theta$ gradually denoises $\bm{z}_T \sim \mathcal{N}(0,1)$ using various samplers~\cite{song2020denoising, liu2022pseudo, lu2022dpm}.

For conditional generation, classifier-free guidance~\cite{ho2022classifier} are employed to jointly train conditional models $\boldsymbol\epsilon_\theta(\bm{z}_t, \bm{c}, t)$ and unconditional models $\boldsymbol\epsilon_\theta(\bm{z}_t, \varnothing, t)$ by dropping out $\bm{c}$, thereby the noise predicted during inference is:
\begin{equation}\label{eq:class_free}
    \boldsymbol\hat{\epsilon}_\theta(\bm{z}_t, \bm{c}, t) = \omega \cdot \boldsymbol\epsilon_\theta(\bm{z}_t, \bm{c}, t) + (1-\omega) \cdot \boldsymbol\epsilon_\theta(\bm{z}_t, \varnothing, t),
\end{equation}
here $\omega$ is the guidance scale which controls the strength of condition $\bm{c}$ in generated results.

\subsection{Paired Concept-Disentangled Dataset}
\label{sec:dataset}
Previous visual concept generation methods~\cite{ruiz2023dreambooth,gal2022image,yang2023paint, wei2023elite, chen2023anydoor} typically use L2 loss to reconstruct training samples, which makes it difficult to extract different concepts from a image. 
Therefore, we design a data construction pipeline to generate paired data specific to certain concepts, named paired concept-disentangled dataset (PCD-200K) with 200K paired data. We divide the visual concepts in the image into three components: content, style, and layout, which are the fundamental visual concepts of an image. Our goal is to describe the desired visual concepts in natural language. Therefore, each pair of our dataset include reference image $\bm{I}_{ref}$ and target image $\bm{I}_{tar}$, reference prompt $\bm{T}_{ref}$ and target prompt $\bm{T}_{tar}$, and concept guide $\bm{T}_{cg}$ in simple natural language, which indicates the shared visual concept between the two images.
Specifically, we design three different pipelines to build different visual concepts. We first apply GPT-4~\cite{achiam2023gpt} to generate reference and target prompts and concept guidance. For content concept, we use FLUX and Kolors-inpainting~\cite{kolors} to generate two images that are different except for the shared subject. For style concept, we apply Instant-Style~\cite{wang2024instantstyle} to generate two images with the same style. For layout, we use ControlNet-Depth~\cite{zhang2023adding} to generate two images with the same layout. Concept guidance for content is the name of the subject (\textit{e.g.} ``chef'', ``lion''), while for the other two dimension is ``style'' and ``layout'', as it is challenging to accurately describe these concepts in natural language.
We give a more detailed description of the data construction pipeline in Sec. II in the supplementary material.

% \subsection{Disentangling Visual Concept For Generation}
\subsection{Language-driven Concept Disentanglement}
\label{sec:cdip}
Images contain rich visual concept representations, and visual encoders like ViT~\cite{dosovitskiy2020image} cannot distinguish these concepts. To efficiently obtain disentangled visual concept representations for generation, we use natural language as guidance to disentangle corresponding concepts from reference image. Previously, Blip-Diffusion~\cite{li2024blip}, DEADiff~\cite{qi2024deadiff}, SSR-Encoder~\cite{zhang2024ssr} and MS-Diffusion~\cite{wang2025msdiffusion} use multi-modal encoders as concept extractor. However, they are all struggled in disentangling general visual concepts from images. Inspired by them, we introduce a pre-trained Q-Former~\cite{li2023blip} with strong multi-modal alignment capabilities as concept extractor to learn from paired data how to disentangle different concepts guided by language, as illustrated in~\cref{fig:framework}.

Specifically, for each pair of our PCD-200K, we utilize a multimodal concept extractor $\bm{E}$ to extract the target concepts from the reference images $\bm{I}_{ref}$ based on the concept guidance $\bm{T}_{cg}$.
The CLIP~\cite{radford2021learning} representations of the reference image $\bm{f}_{I}$ and concept guidance $\bm{f}_{cg}$ are fed into the concept extractor $\bm{E}$ and interact with a learnable query $\bm{q}$ through several cross-attention layers, the output $\bm{H}^{1}_{ce}$ of the first attention layer is: 
\begin{equation}\label{eq:extractor}
\begin{split}
&\bm{H}^{1}_{ce} = \mathrm{Attn}(\mathrm{cat}(\bm{q} + \bm{e}_i, \bm{f}_{cg})\bm{W}^{e}_q, \bm{f}_{I}\bm{W}^{e}_k, \bm{f}_{I}\bm{W}^{e}_v), \\
&{\rm where} \; \mathrm{Attn}(\bm{Q},\bm{K},\bm{V}) = {\rm softmax} (\frac{\bm{Q}\bm{W}^T}{\sqrt{d}})\bm{V},
\end{split}
\end{equation}
where $\bm{e}_i$ is the block embedding corresponds to the i-th diffusion block, $\bm{W}^{e}$ are attention weights of $\bm{E}$. Subsequently, the disentangled concept representation $\bm{f}^i_{cpt}$ output from $\bm{E}$ are fed into the i-th diffusion block through a set of additional cross-attention adapters:
\begin{equation}\label{eq:adapter}
\begin{split}
\bm{H}^{i+1} &= \mu \cdot \mathrm{Attn}(\bm{{H}^i}W_q, \bm{f}^i_{cpt}\bm{W}'_k, \bm{f}^i_{cpt}\bm{W}'_v)\\&+ \mathrm{Attn}(\bm{{H}^i}W_q, \bm{c}_t\bm{W}_k, \bm{c}_t\bm{W}_v), 
\end{split}
\end{equation}
where $\bm{W'}$ are the newly added attention weights, $\bm{W}$ are the original attention weights, $\mu$ is the concept scale of attention output of $\bm{f}^i_{cpt}$, it can be dynamically adjusted during inference to control the concept injection, while being fixed at 1.0 during the training phase. The classifier-free guidance becomes:
\begin{equation}\label{eq:class_free_new}
\begin{split}
\boldsymbol\hat{\epsilon}_\theta(\bm{z}_t, \bm{c}_t, \bm{f}_{cpt}, t) &= \omega \cdot \boldsymbol\epsilon_\theta(\bm{z}_t, \bm{c}_t, \bm{f}_{cpt}, t) \\&+ (1-\omega) \cdot \boldsymbol\epsilon_\theta(\bm{z}_t, \varnothing, \varnothing, t),
\end{split}
\end{equation}

\noindent \textbf{Block Embeddings.}
Recent works~\cite{voynov2024p, qi2024deadiff} have found that different blocks in diffusion model have varying impacts on generated results. Coarse layers tend to learn low-level concepts such as style and color, while fine layers capture high-level semantic concepts. Therefore, some works~\cite{qi2024deadiff, wang2024instantstyle} inject reference style image features into the coarse layers to achieve stylization. However, they rely heavily on manual priors and may ignore the impact of other blocks in results. Agarwa et al.~\cite{agarwal2023image} find that fine layer can also generate appearance-related concepts. To this end, we design a novel block embedding that aims to adapt concept representations to different concept domain of each diffusion block. Specifically, as shown in~\cref{fig:framework} and~\cref{eq:extractor}, we introduce a set of learnable block embeddings $\bm{e}_1$ to $\bm{e}_N$ corresponding to N diffusion blocks. While calculating the $\bm{f}_{cpt}$, each $\bm{e}_i$ is added to $\bm{q}$ and obtain $\bm{f}^i_{cpt}$ that matches different concept domain. These representations are then fed to the corresponding diffusion blocks to interact with latent representations $\bm{H}^i$, achieving adaptive concept domain alignment and controllable concept disentangled generation.

\subsection{Contrastive Orthogonal Disentangled Learning}
\label{sec:contrastive}
Previous works~\cite{li2024blip,qi2024deadiff,zhang2024ssr} use multimodal encoders to extract visual representations guided by language, but they lack a effective concept disentangling mechanism and get conflict results. Inspired by the transferable neighborhood of TraND~\cite{zheng2021trand}, we design a Contrastive Orthogonal Disentangled (COD) learning mechanisms in training stage to disentangle each types of visual concepts into mutually orthogonal dimensions in the representation space. Since these representations are orthogonal, they can be easily combined for multi-concept generation without conflict. 

Specifically, we design a learnable anti-query $\bm{q}_a$ of the same size as $\bm{q}$ to capture irrelevant concepts $\bm{f}^a_{cpt}$ in the reference image as~\cref{eq:extractor}. To learn $\bm{q}$ and $\bm{q}_a$, we input $\bm{I}_{tar}, \bm{q}$ and the concept guidance $\bm{T}_{cg}$ into our concept extractor to obtain the concept representation $\bm{f}^{tar}_{cpt}$ in $\bm{I}_{tar}$. Subsequently, $\bm{f}^a_{cpt}$ and $\bm{f}^{tar}_{cpt}$ are concatenated and fed to the additional cross-attention layers of U-Net to generate $\bm{I}_{ref}$. This additional training branch facilitates the combine generation of different concepts. Furthermore, we design a contrastive orthogonal disentangled Loss $\mathcal{L}_{COD}$, which constrains $\bm{f}_{cpt}$ and $\bm{f}^{tar}_{cpt}$ to be similar in the representation space, and $\bm{f}^a_{cpt}$ and $\bm{f}^{tar}_{cpt}$ to be orthogonal:
\begin{equation}\label{eq:COD}
\mathcal{L}_{COD} = \rvert\bm{cos}(\bm{f}^a_{cpt},\bm{f}^{tar}_{cpt})\rvert - \bm{cos}(\bm{f}_{cpt},\bm{f}^{tar}_{cpt}),
\end{equation}
where $\bm{cos}(\cdot)$ is cosine similarity, and the total loss is:
\begin{equation}\label{eq:loss}
\mathcal{L}_{total} = \mathcal{L}_{ldm} + \lambda \cdot \mathcal{L}_{COD},
\end{equation}
where $\lambda$ is the weight to control the influence of $\mathcal{L}_{COD}$.

\section{Experiments}
\label{sec:exp}

\begin{figure*}[ht]
    \centering
    \includegraphics[width=0.95\textwidth]{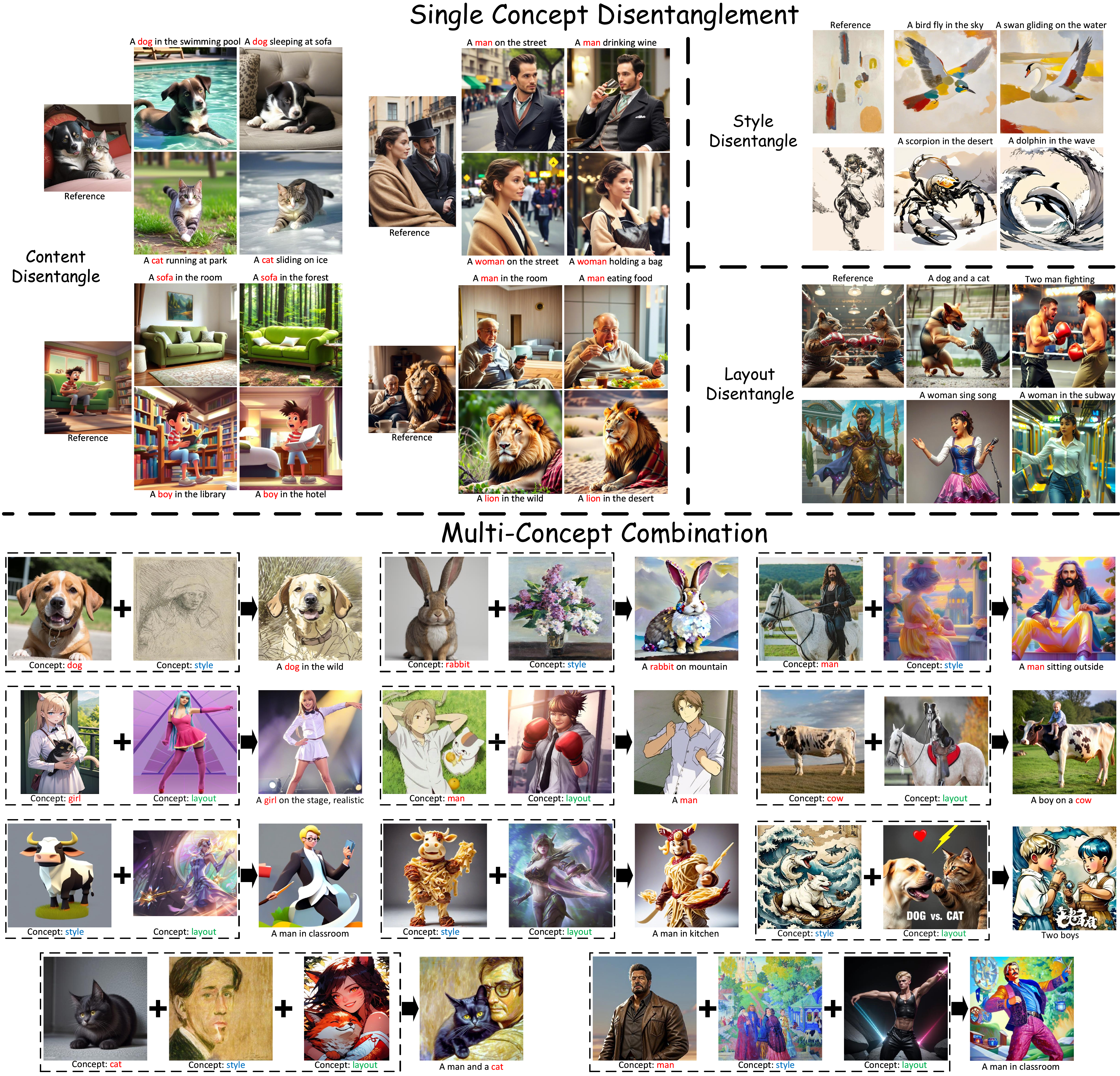}
    \caption{%
    \textbf{Diverse capabilities of our method.} Our method supports the single concept disentangled generation from a same reference image, including different content, style, and layout. In addition, we can combine these disentangled concepts to generate results that incorporate multiple desired concepts.
    }
    \label{fig:case}
    \vspace{-10pt}
\end{figure*}

\subsection{Experimental Settings}
\label{setting}

\noindent \textbf{Datasets.} OmniPrism is trained on our PCD-200K dataset (\cref{sec:dataset}) with 200K paired data, each pair has two distinct images sharing a same concept. For evaluation, we assesses single-subject concept generation performance on DreamBench~\cite{ruiz2023dreambooth}, which contains 30 subjects and 25 prompts, yielding 3,000 samples (4 random samples per combination). For systematically evaluating concept-disentangled performance, we build a DisenBench with 40 dual-subject images,  60 diverse style images, and 20 human figure images, yielding 640 samples (4 random samples per input).

\noindent \textbf{Evaluation Metrics.}
Following previous works, we measure the CLIP-I~\cite{radford2021learning} and DINO~\cite{zhang2023dino} for subject consistency, and CLIP-T~\cite{jack2021clipscore} for text fidelity on DreamBench. To evaluate the concept disentanglement performance on DisenBench, we adopt Mask CLIP-I and Mask DINO for specific content consistency. CLIP-T for text fidelity. Style Similarity~\cite{somepalli2024measuring} to measure the style similarity. Salient structure distance (SSD)~\cite{park2012modeling} for layout consistency. Aesthetic Score~\cite{christoph2022laion} for image quality. The detail of all metrics is described in the Sec. III of supplementary material.

\noindent \textbf{Compared Methods.}
To fully evaluate our superiority, we compare our method with the SOTA visual concept generation methods, including IP-Adapter~\cite{ye2023ip}, BLIP-Diffusion~\cite{li2024blip}, DEADiff~\cite{qi2024deadiff}, SSR-Encoder~\cite{zhang2024ssr}, MS-Diffusion~\cite{wang2025msdiffusion}, OmniGen~\cite{xiao2024omnigen}. Both these methods and our implementation are described in detail in Sec. III of supplementary material.

\noindent \textbf{Implementation Details. }
We use Stable Diffusion XL (SDXL) \cite{podell2023sdxl} as the base model, add an extra cross-attention layer to U-Net as IP-Adapter\cite{ye2023ip}, and initialize it with the weights of pretrained IP-Adapter-XL-Plus \cite{ye2023ip}. We use the pretrained BLIP2 \cite{li2023blip} Q-Former as the concept extractor, and add a project layer to project its output to the input size of U-Net's cross-attention. The image input and concept guidance are fed into the concept extractor after the image encoder and text encoder of CLIP-L/14 \cite{radford2021learning} respectively. The token number of learnable query, anti-query and block embeddings are set to 32. 
During training, we freeze other parameters and only train the concept extractor with its project layer, and the extra cross-attention layer in U-Net. 
Our model is trained on a single machine with 8*80GB A100 GPUs by AdamW optimizer \cite{loshchilov2018decoupled}. We adopt a three-stage training method. The first two stages refer to IP-Adapter-SDXL, we train our model for 15,000 steps and 10,000 steps on the Laion Aesthetics 6.5+ dataset\footnote{https://laion.ai/blog/laion-aesthetics/} at 512 and 1024 resolutions respectively. $\mathcal{L}_{COD}$ is not used in these two stages, and the learning rate of all trainable parameters is 1$\bm{e}$-5. In the third stage, the model is trained on our PCD-200K with $\mathcal{L}_{total}$ for 80000 steps, with a batch size of 4 per GPU, the learning rate of the extra cross-attention layer is 1$\bm{e}$-6, and the learning rate of others is 1$\bm{e}$-5. 
During inference, we adopt DDIM sampler~\cite{song2020denoising} with 20 sample steps and the guidance scale of classifier-free guidance $\omega$ is 5.

\begin{table*}[t]

    \caption{\textbf{Quantitative comparison with other methods.} The 1st and 2nd ranks for each metric are marked in \textbf{bold} and \underline{underline}.}
    \centering
    \scalebox{0.95}{
        \begin{tabular}{@{}l|ccc|cccccc@{}}
        \toprule
        \multirow{2}{*}{Method}   & \multicolumn{3}{c}{DreamBench} & \multicolumn{6}{c}{DisenBench} \\ \cmidrule(r){2-4} \cmidrule(r){5-10}
                                    & CLIP-I & DINO & CLIP-T      & \makecell[c]{Mask\\CLIP-I} $\uparrow$ & \makecell[c]{Mask\\DINO} $\uparrow$   & CLIP-T $\uparrow$  & \makecell[c]{Style\\Similarity} $\uparrow$ & SSD $\downarrow$ & \makecell[c]{Aesthetic\\Score} $\uparrow$       \\ \hline
        IP-Adapter \cite{ye2023ip}  & 0.809 & 0.608  & 0.274   & 0.784    & \underline{0.506}       & 0.243            & \textbf{0.804} & \underline{0.268}  & 6.185                \\
        BLIP-Diffusion \cite{li2024blip} & 0.779 & 0.594 & 0.300 & 0.755   & 0.488     & 0.249              & 0.512     & 0.369     & 6.174                 \\
        DEADiff \cite{qi2024deadiff}  & 0.770 & 0.532  & 0.306      & 0.736   & 0.379      & 0.273             & 0.349     & 0.408    & 6.110                  \\
        SSR-Encoder \cite{zhang2024ssr} & \textbf{0.821} & 0.612  & 0.308   & \underline{0.793}   & 0.504      & 0.259              & 0.583   & 0.345     & 6.370        \\ 
        MS-Diffusion \cite{wang2025msdiffusion}  & 0.792 & \textbf{0.671} & \underline{0.321}   & 0.750   & 0.443     & 0.285              & 0.326   & 0.324     & 6.403         \\ 
        OmniGen \cite{xiao2024omnigen} & 0.801 & 0.570  & 0.315   & 0.776   & 0.480    & \underline{0.287}              & 0.339   & 0.294     & \underline{6.410}         \\ 
        Ours      & \underline{0.815} & \underline{0.614}  & \textbf{0.328}     & \textbf{0.797} & \textbf{0.513} & \textbf{0.296}     & \underline{0.585}   & \textbf{0.267}     & \textbf{6.485}      \\ \bottomrule
        \end{tabular}
    }
    % \vspace{-1mm}
    
    \label{table:quan}
    % \vspace{-2mm}
\end{table*}

\begin{figure*}[ht]
    \centering
    \includegraphics[width=0.95\textwidth]{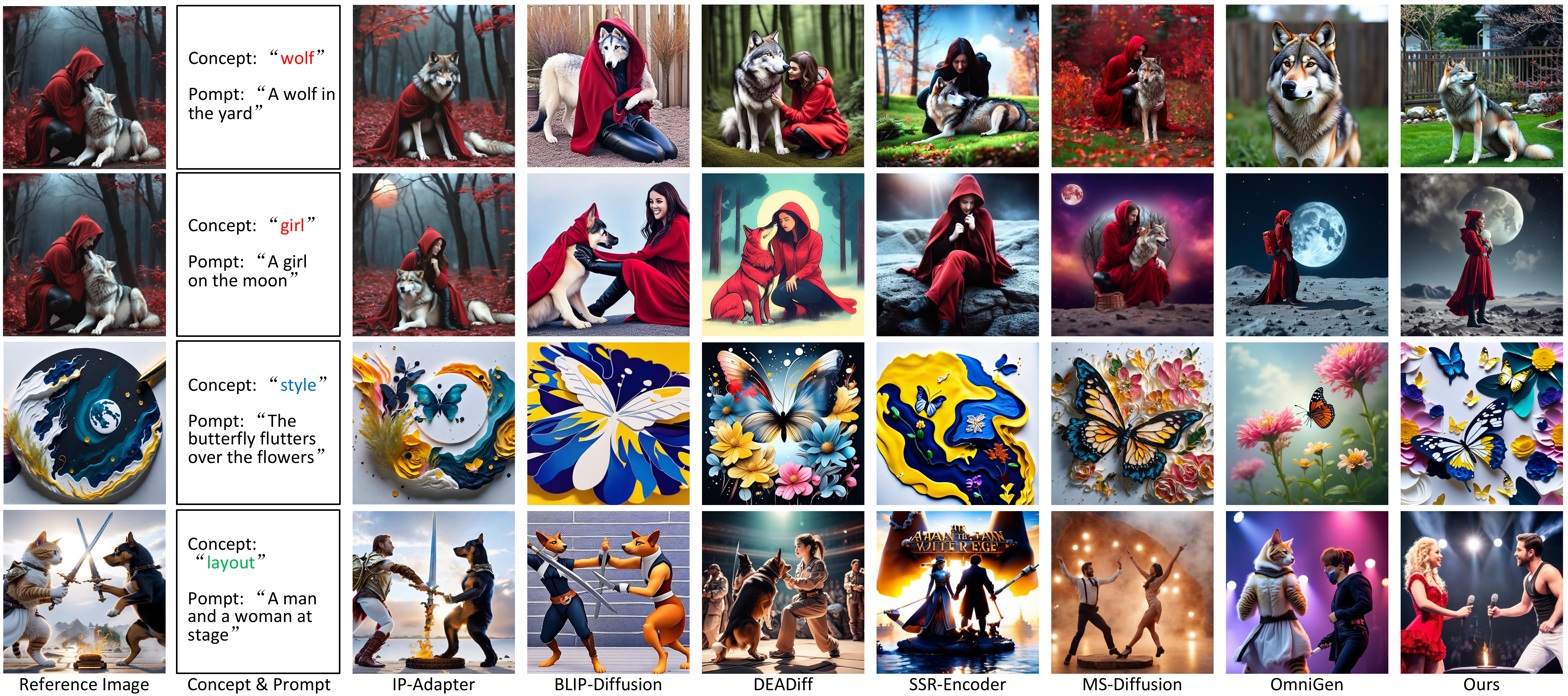}
    \caption{%
    \textbf{Comparison with the state-of-the-art works.} Our method achieves superior disentangled generation performance. It not only avoids introducing irrelevant concepts but also ensures the highest concept consistency, text fidelity and image quality.
    }
    \label{fig:compare}
\end{figure*}

\subsection{Main Results}
\noindent \textbf{Quantitative Evaluation. }
% \subsection{Quantitative Evaluation}
\cref{table:quan} shows the quantitative comparison results between our method and others. For single-subject concept generation, we achieve the 2nd subject consistency and the 1st text fidelity. For concept disentanglement, our method achieves the highest Mask CLIP-I / DINO and CLIP-T scores, which indicates our superior disentangled concept consistency and prompt fidelity. IP-adapter achieves the highest style similarity but at the cost of the lowest text faithfulness, which indicates that they are overly reliant on reference images and neglect the text prompt. Our method achieves the highest style similarity except for IP-Adapter. In addition, our layout consistency and image quality are significantly better than other methods. In summary, our method achieves optimal results in terms of text fidelity, concept consistency, and image quality, while maintain the performance on subject consistency in single-subject concept generation. We further evaluate the disentanglement scores on concept representations in Sec. IV.C of the supplementary material.

% \subsection{Qualitative Evaluation}
\noindent \textbf{Qualitative Evaluation. }
We demonstrate the capabilities of our method from multiple aspects, as shown in~\cref{fig:case}. Our method supports the disentangled generation of multiple concepts guided by natural language, including single subject, individual subjects within multiple subjects, style, and layout (\textit{e.g.} relation, pose). These results proves that our method possesses a robust concept disentangled representation space and a strong visual concept disentangled generation capability. In addition, by constraining different types of concept representations to be orthogonal, these features can be naturally integrated into the same generated result in any combination without interfering with each other. We further add more extra visual results in supplementary images.

The comparison results are shown in~\cref{fig:compare}. It can be seen that existing visual concept generation methods struggled to disentangle different concepts from reference images and leading to concept leakage. IP-Adapter and BLIP-Diffusion are too faithful to the reference image and generate concepts irrelevant to prompts. DEADiff relies on specific diffusion blocks to disentangle style, but it cannot disentangle content and layout. SSR-Encoder aligns image representations to the concepts in CLIP text domain, but they suffer from the trade-off of either inaccurately generating concepts or generating irrelevant concepts. MS-Diffusion and OmniGen have poor results in style and layout disentanglement, and generate irrelevant concepts sometimes. In contrast, our method effectively disentangle different concept representations based on concept guidance and aligns them to the concept domains of corresponding diffusion blocks, thus achieving superior concept fidelity and concept independence across multiple concepts.

\subsection{Visualization of Concept Disentanglement}

\begin{figure*}[t]
    \centering
    \includegraphics[width=0.98\textwidth]{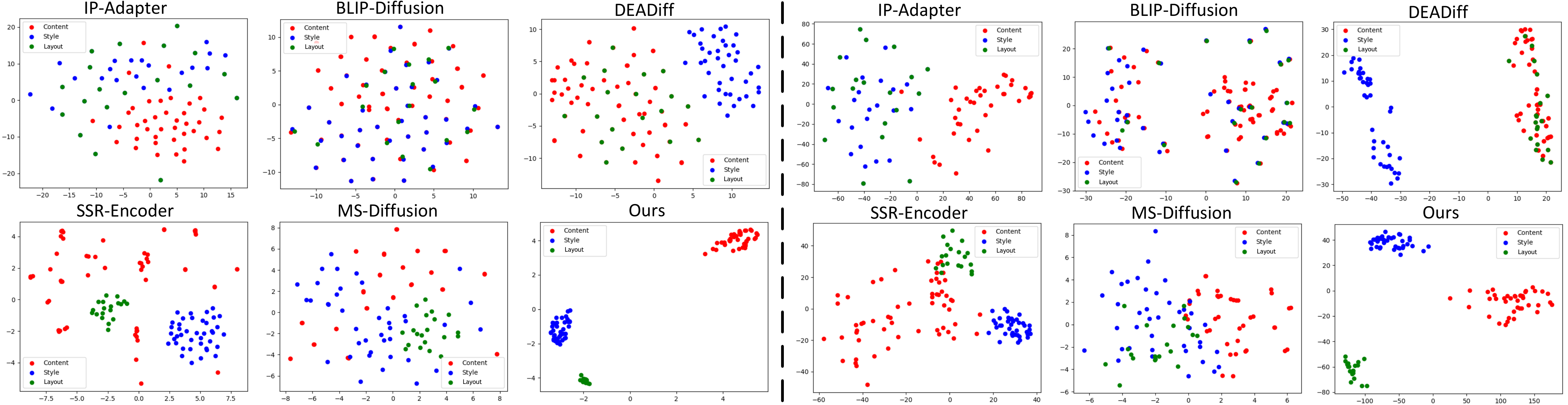}
    \caption{%
    \textbf{The t-SNE projection (left) and PCA (right) visualization of concept representations with other methods.}
    }
    \label{fig:tsne_pca}
    \vspace{-10pt}
\end{figure*}

\noindent \textbf{Concept Representations Distance. }
Our \textbf{OmniPrism} is designed to learn and disentangle concepts with varying semantics. We expect that these concepts be distributed across distinct clusters, so we calculate the t-SNE projection~\cite{van2008visualizing} and Principal component analysis (PCA)~\cite{abdi2010principal} of different concept representations for all compared methods. We do not visualize the result of OmniGen since they have no learnable concept representations. As shown in~\cref{fig:tsne_pca}, all compared methods failed to disentangle different concepts, SSR-Encoder can separate style and layout concepts since these terms are distinct in the text space, but it struggles to distinguish content concepts with complex semantics. In contrast, our \textbf{OmniPrism} disentangles each concept into separate clusters, demonstrating our superior concept disentangling ability. 

\begin{figure}[t]
    \centering
    \includegraphics[width=0.48\textwidth]{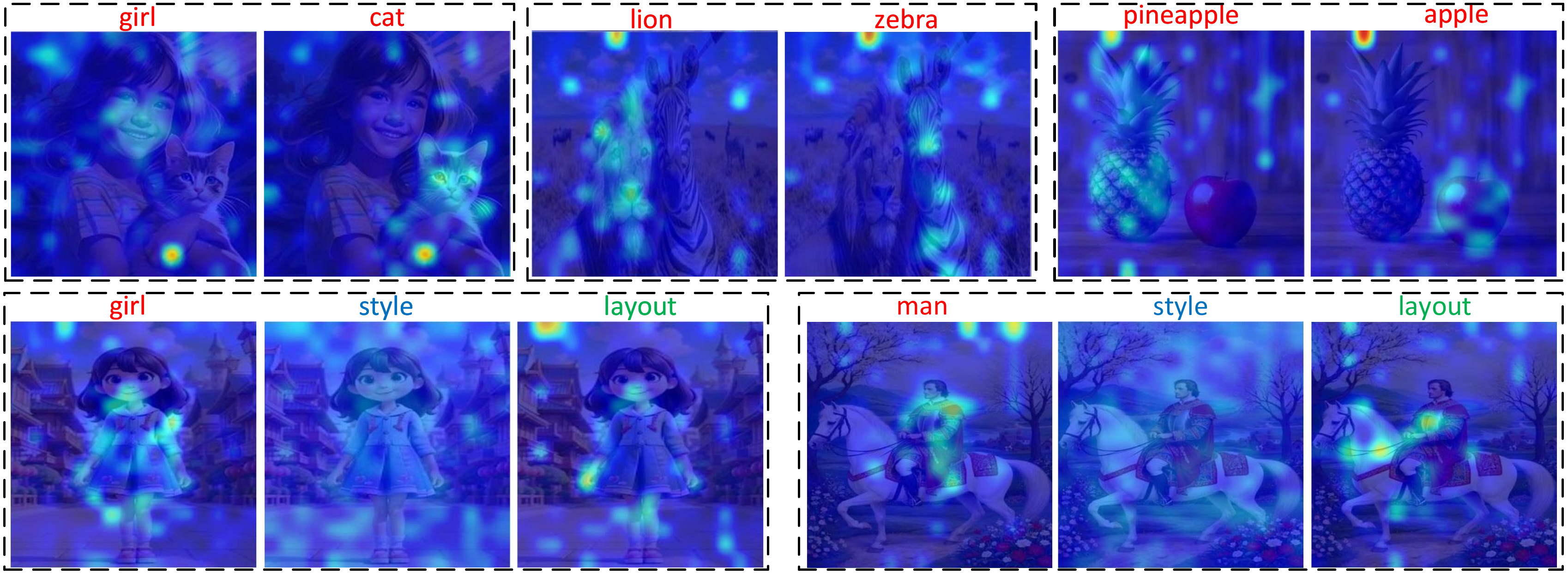}
    \caption{%
    \textbf{Visualization of attention map.} We show how concept guidance interacts with image representations in concept extractor.
    }
    \label{fig:attn}
    \vspace{-10pt}
\end{figure}

\noindent \textbf{Concept Representations Attention Map. }
To demonstrate that our concept representations are disentangled, we visualize the attention maps of concept guidance alongside reference image representations within the concept extractor, as shown in~\cref{fig:attn}. Different concept guidance aligns with the corresponding visual concepts in the reference image, thus showing our precise language-driven disentanglement capability.

\subsection{Discussion with ControlNet on Layout}
\begin{figure}[t]
    \centering
    %\vspace{-10mm}
    \includegraphics[width=0.47\textwidth]{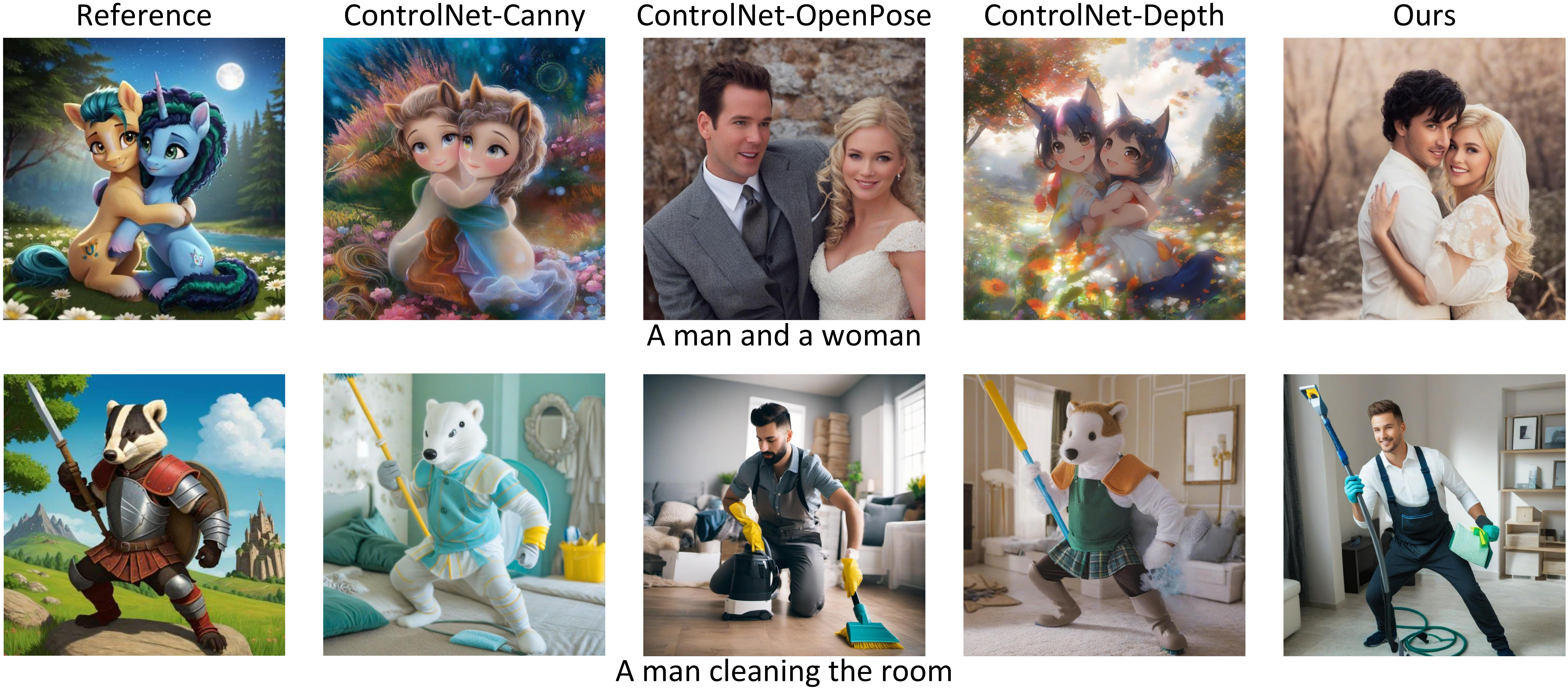}
    %\vspace{-5pt}
    \caption{%
    \textbf{Discussion with ControlNet.} ControlNet with all conditions is prone to conflicts between prompts and structural features, while our method extracts abstract ``layout" concepts (\textit{e.g.} relationships, poses) and generates creative results.
    }
    \label{fig:dis_ctrl}
    \vspace{-10pt}
\end{figure}

To train our method to learn the ``layout" concepts, we use data generated from ControlNet-Depth \cite{zhang2023adding}. However, the ``layout" concept generation ability of our \textbf{OmniPrism} is different from ControlNet-Depth. ControlNet concatenate the depth map to the pixel space, which is a strict constraint and may causing conflict between prompt and depth map. As shown in \cref{fig:dis_ctrl},  whether ControlNet uses canny, openpose, or depth as the control condition, the prompt may conflict with the control condition, either failing to generate the corresponding structure accurately or not following the prompt guidance. In contrast, our layout concept is injected into the latent space via cross-attention, enabling flexible generation of concepts such as relationships without conflicts between prompts and structural features. 

We also measure the SSD and CLIP-T scores of the three ControlNet in \cref{table:ctrl}. As the results in \cref{fig:dis_ctrl}, ControlNet-Canny and ControlNet-Depth are consistent with the layout of the reference image, but have poor text alignment when prompt is conflict with reference image. ControlNet-Openpose has a high text alignment, yet it ignores the layout of the reference image. In contrast, our method achieves a better trade-off between these two metrics and is a spatial layout disentangling method with a different application scenario from ControlNet.

\begin{figure}[t]
    \centering
    \includegraphics[width=0.47\textwidth]{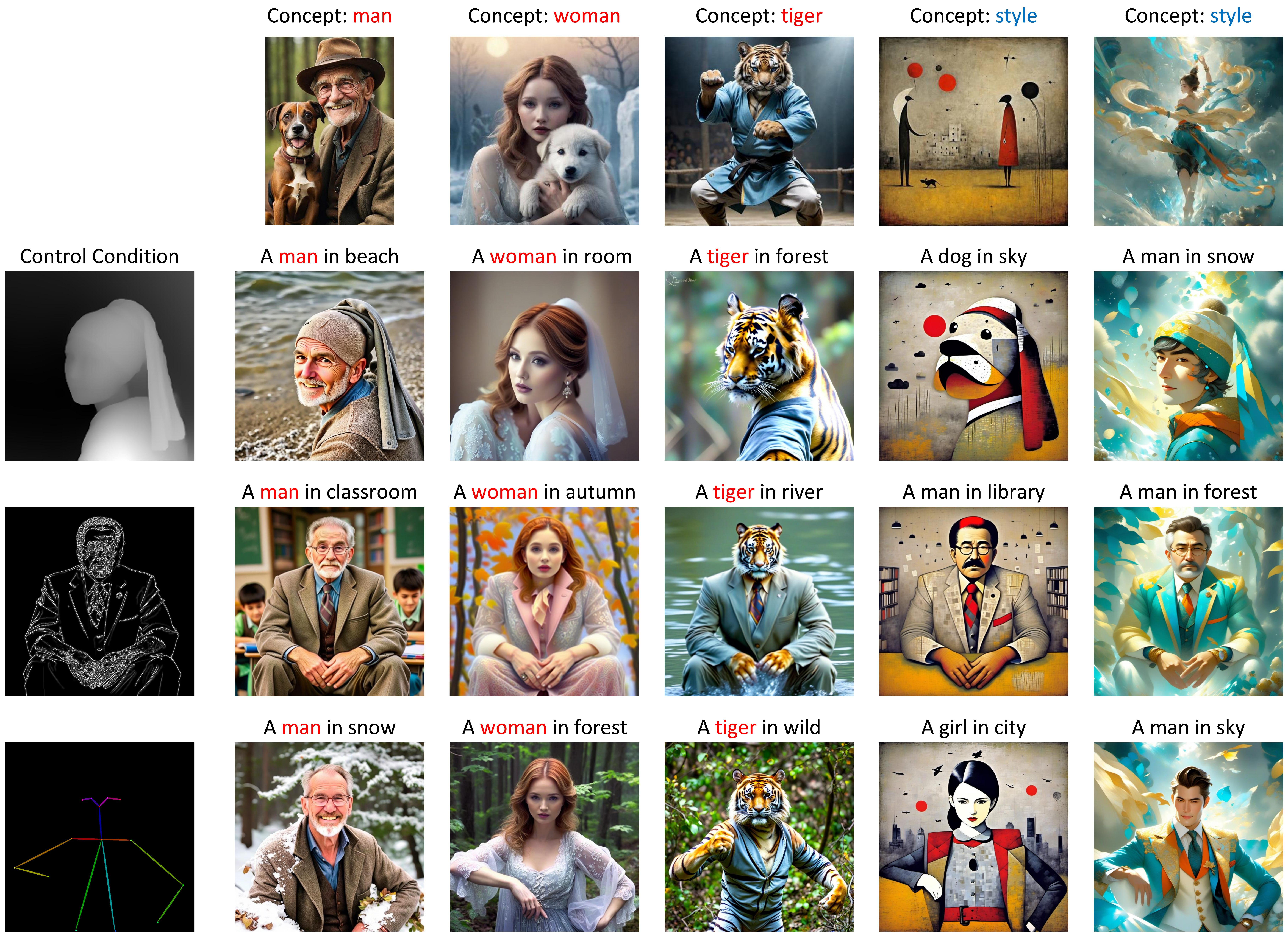}
    \caption{%
        \textbf{Additional Controls with ControlNet.}
    }
    \label{fig:ctrl_omni}
    % \vspace{-5pt}
\end{figure}

Our layout concept generation is not mutually exclusive with ControlNet, but exists as two complementary control methods. We also support additional control with ControlNet, as shown in \cref{fig:ctrl_omni}. It can be seen that our \textbf{OmniPrism} naturally supports various categories of ControlNet's control over content and style concepts. 

\begin{table}[t]
\centering
\caption{\textbf{Evaluation of Layout.}}
\label{table:ctrl}
\scalebox{0.95}{
\begin{tabular}{@{}lcccc@{}}
\toprule
Method          & \makecell[c]{ControlNet\\Canny}  & \makecell[c]{ControlNet\\OpenPose}  & \makecell[c]{ControlNet\\Depth} & Ours       \\ \midrule
SSD $\downarrow$            & \textbf{0.169} & 0.290     & \underline{0.175}       & 0.267 \\
CLIP-T $\uparrow$            & 0.236 & \textbf{0.296}     & \underline{0.256}       & \textbf{0.296} \\ \bottomrule
\end{tabular}
}
% \vspace{-13pt}

\end{table}

\subsection{Ablation Study}
\begin{figure}[t]
    \centering
    \includegraphics[width=0.48\textwidth]{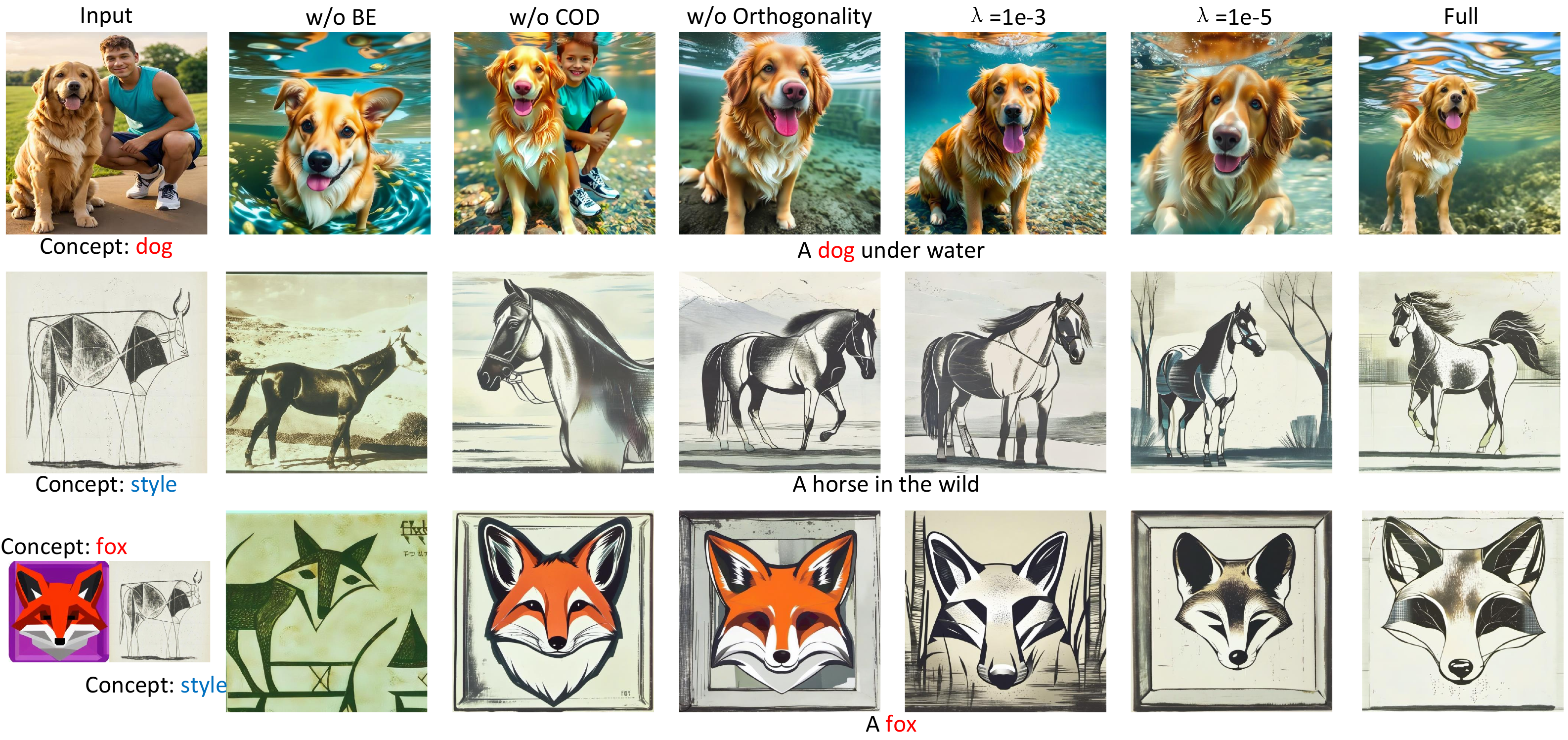}
    \caption{%
    \textbf{Ablation of each component.} We evaluate removing block embedding (BE), contrastive orthogonal disentangled (COD) learning, and orthogonality from our method respectively.
    }
    \label{fig:abla}
\end{figure}

To evaluate the the disentangling effectiveness of each component in our method, we conduct ablation experiments by removing each component respectively. For block embeddings (BE), we simply remove them. For contrastive orthogonal disentangled (COD) learning, we firstly remove the orthogonality (Orth) by modifying the~\cref{eq:COD} to:
\begin{equation}\label{eq:COD_abla}
\mathcal{L}_{CD} = \bm{cos}(\bm{f}^a_{cpt},\bm{f}^{tar}_{cpt}) - \bm{cos}(\bm{f}_{cpt},\bm{f}^{tar}_{cpt}),
\end{equation}
then we remove all the contrastive orthogonal disentangled learning by delete the anti-query $\bm{q}_a$ and $\mathcal{L}_{COD}$. The qualitative results are shown in~\cref{fig:abla}. And the quantitative results are shown in~\cref{table:abla}. While orthogonality slightly reduces disentangled subject consistency, it enhances disentanglement between different concepts, as improved style similarity in~\cref{fig:abla}.

\noindent \textbf{Effect of Block Embedding. }Our block embedding is designed to adaptively match the concept representations to the concept domains of different diffusion blocks. As shown in the second column of~\cref{fig:abla}, after removing the block embedding, the extracted concepts are not well aligned to the concept domain of the diffusion model, the generated results exhibit discrepancies with the concepts in the reference image.

% \begin{table*}[t]
%     \scalebox{0.85}{
%         \begin{tabular}{@{}l|ccc|cccccc@{}}
%         \toprule
%         \multirow{2}{*}{Method}   & \multicolumn{3}{c}{DreamBench} & \multicolumn{6}{c}{DisenBench} \\ \cmidrule(r){2-4} \cmidrule(r){5-10}
%                                     & CLIP-I & DINO & CLIP-T      & \makecell[c]{Mask\\CLIP-I} $\uparrow$ & \makecell[c]{Mask\\DINO} $\uparrow$   & CLIP-T $\uparrow$  & \makecell[c]{Style\\Similarity} $\uparrow$ & SSD $\downarrow$ & \makecell[c]{Aesthetic\\Score} $\uparrow$       \\ \midrule
%         w/o BE  & 0.795 & 0.562  & 0.313  & 0.782   & 0.479      & 0.279              & 0.501   & 0.298     &  6.141        \\ 
%         w/o Contrastive  & \underline{0.803} & 0.574 & 0.320   & 0.781   & 0.507   & 0.286              & 0.500   & 0.293     & \underline{6.378}         \\ 
%         w/o Orthogonality & 0.800 & \underline{0.596}  & \underline{0.322}   & \textbf{0.801}    & \textbf{0.517}   & \underline{0.288}              & \underline{0.550}    & \underline{0.282}     & 6.343          \\ 
%         Full Method      & \textbf{0.815} & \textbf{0.614}  & \textbf{0.328}     & \underline{0.797} & \underline{0.513} & \textbf{0.296}     & \textbf{0.585}   & \textbf{0.267}     & \textbf{6.485}      \\\bottomrule
%         \end{tabular}
%     }
%     \vspace{-1mm}
%     \caption{\textbf{Evaluation of Ablation Study.} The 1st and 2nd ranks for each metric are marked in bold and underlined.}
%     \label{table:abla}
%     \vspace{-2mm}
% \end{table*}

\begin{table}[t]
    \caption{\textbf{Ablation Study of OmniPrism.}}
    \scalebox{0.82}{
        \begin{tabular}{@{}lcccccc@{}}
        \toprule
        Method   
                                    & \makecell[c]{Mask\\CLIP-I} $\uparrow$ & \makecell[c]{Mask\\DINO} $\uparrow$   &  CLIP-T $\uparrow$  & \makecell[c]{Style\\Similarity} $\uparrow$ & SSD $\downarrow$  & \makecell[c]{Aesthetic\\Score} $\uparrow$   \\ \midrule
        w/o BE    & 0.782   & 0.479      & 0.279              & 0.501   & 0.298     & 6.141        \\ 
        w/o COD     & 0.781   & 0.507   & 0.286              & 0.500   & 0.293       & 6.378      \\ 
        w/o Orth    & \textbf{0.801}    & \textbf{0.517}   & \underline{0.288}              & 0.550    & 0.282 & 6.343        \\ 
        Full Method        & \underline{0.797} & \underline{0.513} & \textbf{0.296}     & \textbf{0.585}   & \textbf{0.267}  & \textbf{6.485}        \\ \hline
        $\lambda$=1e-3  &0.780   & 0.456      & 0.294             & 0.538    & 0.289  & 6.365    \\
        $\lambda$=1e-5  & 0.786   & 0.511      & 0.286             & \underline{0.559}    & \underline{0.270} & \underline{6.471}  \\ \bottomrule
        \end{tabular}
    }
    % \vspace{-1mm}
    
    \label{table:abla}
    % \vspace{-2mm}
\end{table}

\noindent \textbf{Effect of COD. }Our COD learning mechanism is designed to help the concept disentanglement. As shown in the third column of~\cref{fig:abla}, without it, the generated results contain concepts irrelevant to concept guidance (\textit{e.g.}, the boy in the first row), and the fidelity to the desired concept is compromised by these irrelevant concepts.

\noindent \textbf{Effect of Orthogonality. }Removing only the orthogonality in COD Learning can still produce concept disentangled results. However, when combining multiple concepts, these non-orthogonal concept representations interfere with each other and become mixed, as shown in the 4th column of~\cref{fig:abla}. By adding orthogonality, the interference between concepts is greatly reduced when combining multiple different types of concepts, resulting in the ideal concept combination results, as shown in the last column of~\cref{fig:abla}.

\noindent \textbf{Effect of the weight $\lambda$ of $\mathcal{L}_{COD}$. }
$\lambda$ is set to 1e-4 as default. As shown in~\cref{table:abla}, high $\lambda$ will affect image quality, and low $\lambda$ will affect the disentangling ability.

% \noindent \textbf{Effect of Concept Scale $\mu$ in inference. }
% The concept scale $\mu$ of concept guidance controls the influence level of concepts on the generated results. As shown in \cref{fig:ip_scale}, reducing $\mu$ decreases the similarity between the concepts in the generated results and the reference concepts. When $\mu$ is too large, the image quality will be significantly deteriorates. Notably, even when the scale is excessively high and image quality degrades, irrelevant concepts from the reference image do not appear in the generated results, demonstrating our superior concept disentanglement ability. By adjusting $\mu$, users can generate results that better align with their expectations.

\section{Limitations}
\label{sec:limit}
\begin{figure}[t]
    \centering
    \includegraphics[width=0.47\textwidth]{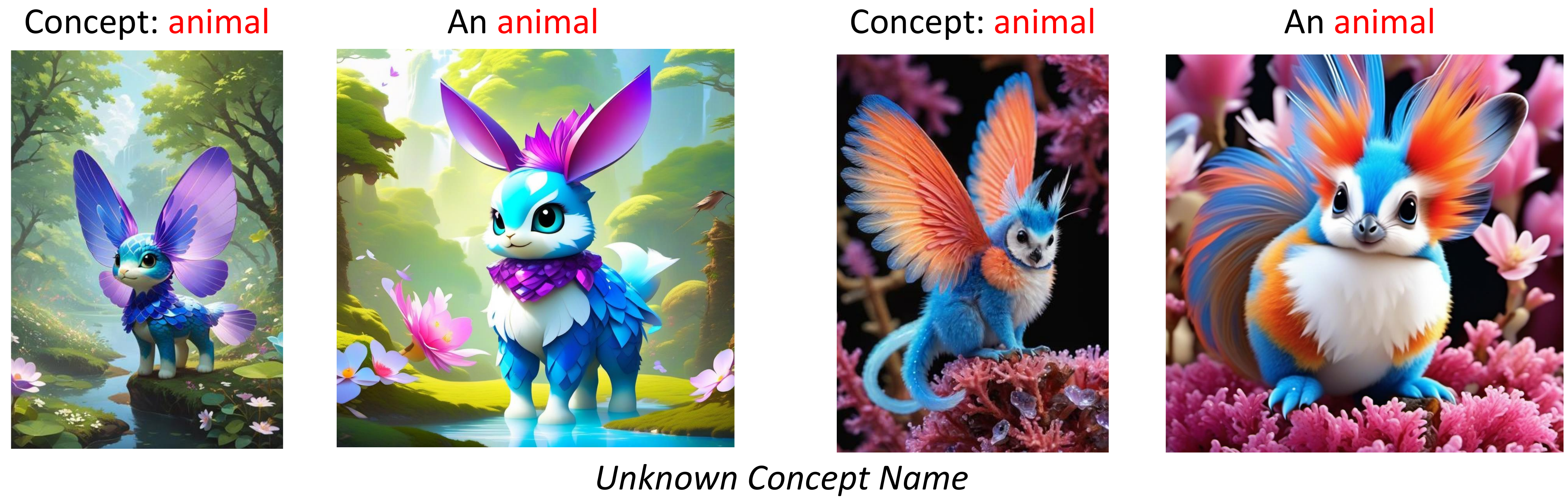}
    \caption{%
        \textbf{Limitations of our OmniPrism.} Our method may fail when the concept name is unknown.
    }
    \label{fig:limit}
    % \vspace{-5pt}
\end{figure}

Our \textbf{OmniPrism} can disentangle and generate various concepts in an image and allowing for any combination in a single image. However, when the concepts are difficult to describe in natural language, such as unknown categories, our method struggles to generate similar concepts. This is because these objects are challenging for the multimodal extractor to capture through natural language, as shown in \cref{fig:limit}. 
In future work, we plan to expand our dataset to encompass a broader and more detailed natural language field to address this issue and achieve more accurate concept disentanglement.

\section{Conclusion}
\label{sec:conclusion}
In this paper, we propose \textbf{OmniPrism}, an innovative approach for disentangled visual concept generation that disentangles individual concepts or combines multiple concepts into outputs.
By integrating a multimodal Q-Former as a concept extractor and a Contrastive Orthogonal Disentangle (COD) learning mechanism, our method effectively disentangles different visual concepts into distinct clusters within the representation space guided by natural language. A novel block embedding further enhances the alignment of the concept representations with the diffusion block's concept domains, allowing for high-fidelity generation that closely matches prompts and desired concepts. Moreover, the construction of the Paired Concept Disentanglement Dataset (PCD-200K) offers a valuable resource for advancing research in this field. 
Our \textbf{OmniPrism} effectively solves the concept leakage problem through COD learning, while supporting flexible combination of multiple concepts across images, providing a new paradigm for creative visual generation. Future work could explore extending this framework to more complex scenarios and further refining the disentanglement process to enhance the diversity and applicability of generated concepts.

\bibliographystyle{IEEEtran}
\bibliography{main}

\clearpage
\renewcommand{\thesection}{\Alph{section}}
% \maketitlesupplementary

\noindent In the supplementary materials, we introduce more detailed analysis and additional results:
\begin{itemize}
\item \cref{sec:sup_explanation} explains the basic \textbf{visual concepts} and shows some other refined concept generation results (\cref{fig:met}).
\item \cref{sec:sup_dataset} demonstrates our \textbf{PCD-200K dataset} (\cref{fig:data_pipe}).
\item \cref{sec:sup_exp} provides \textbf{more experimental details}, including \textit{metric details} and \textit{comparative methods}.
\item \cref{sec:sup_analy} introduces \textbf{more experiments} of our method in \textit{compatibility to base models} (\cref{fig:diff_model}, \cref{table:sup_quan}), \textit{additonal comparisons} (\cref{fig:obj}, \textit{evaluation on disentangling capability} (\cref{table:sup_dci}), \textit{multi-content disentanglement and combinations} (\cref{fig:multi_obj}, \cref{fig:3obj}), and \textit{Results of Complex Prompts.} (\cref{fig:complex}). 
\item \cref{sec:sup_res} shows \textbf{more qualitative results} of disentangled generation of \textit{content} (\cref{fig:sup_obj}), \textit{style} (\cref{fig:sup_style}), and \textit{composition} (\cref{fig:sup_com}).
\item \cref{sec:sup_impact} discusses the \textbf{social impact} of our work.

\end{itemize}

\section{Explanation of Visual Concepts}
\label{sec:sup_explanation}
\begin{figure}[ht]
    \centering
    %\vspace{-10mm}
    \includegraphics[width=0.47\textwidth]{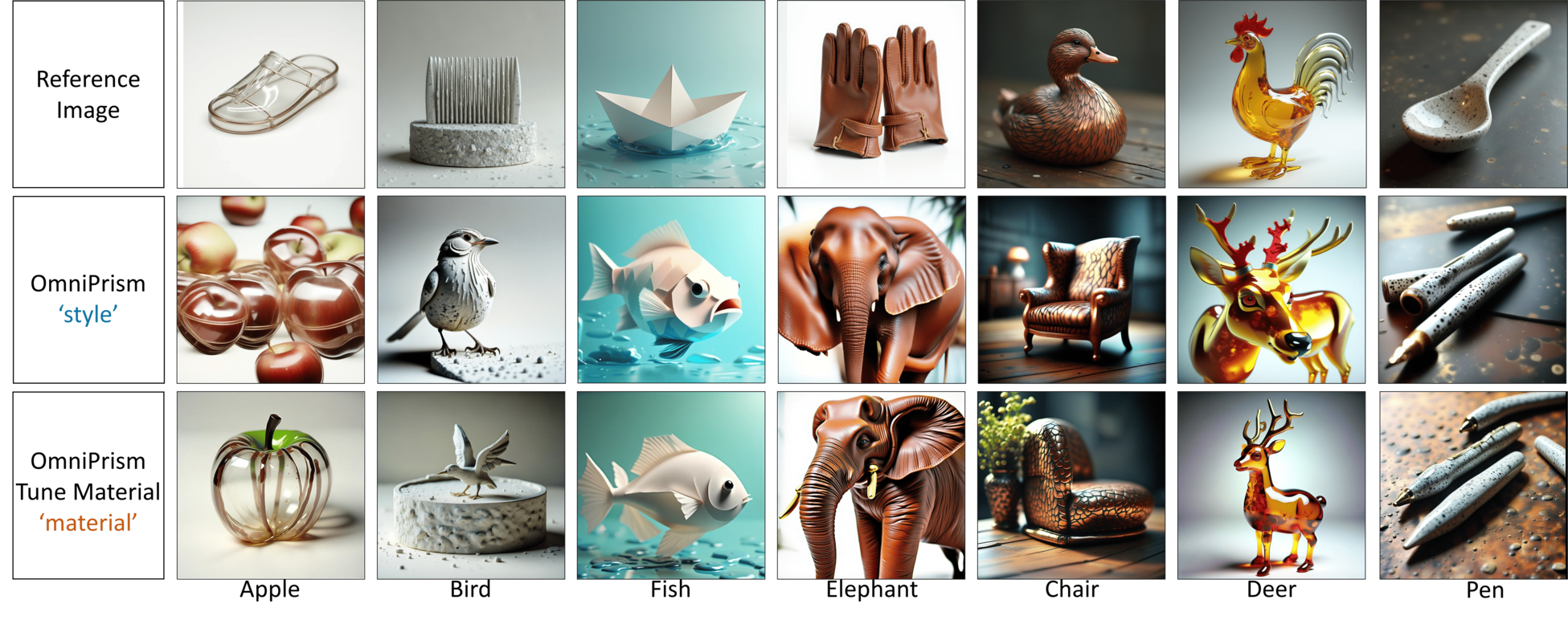}
    \vspace{-5pt}
    \caption{\textbf{Qualitative Result on Material Concepts.}}
    \label{fig:met}
    \vspace{-15pt}
\end{figure}

\begin{figure*}[!t]
    \centering
    \includegraphics[width=1.0\textwidth]{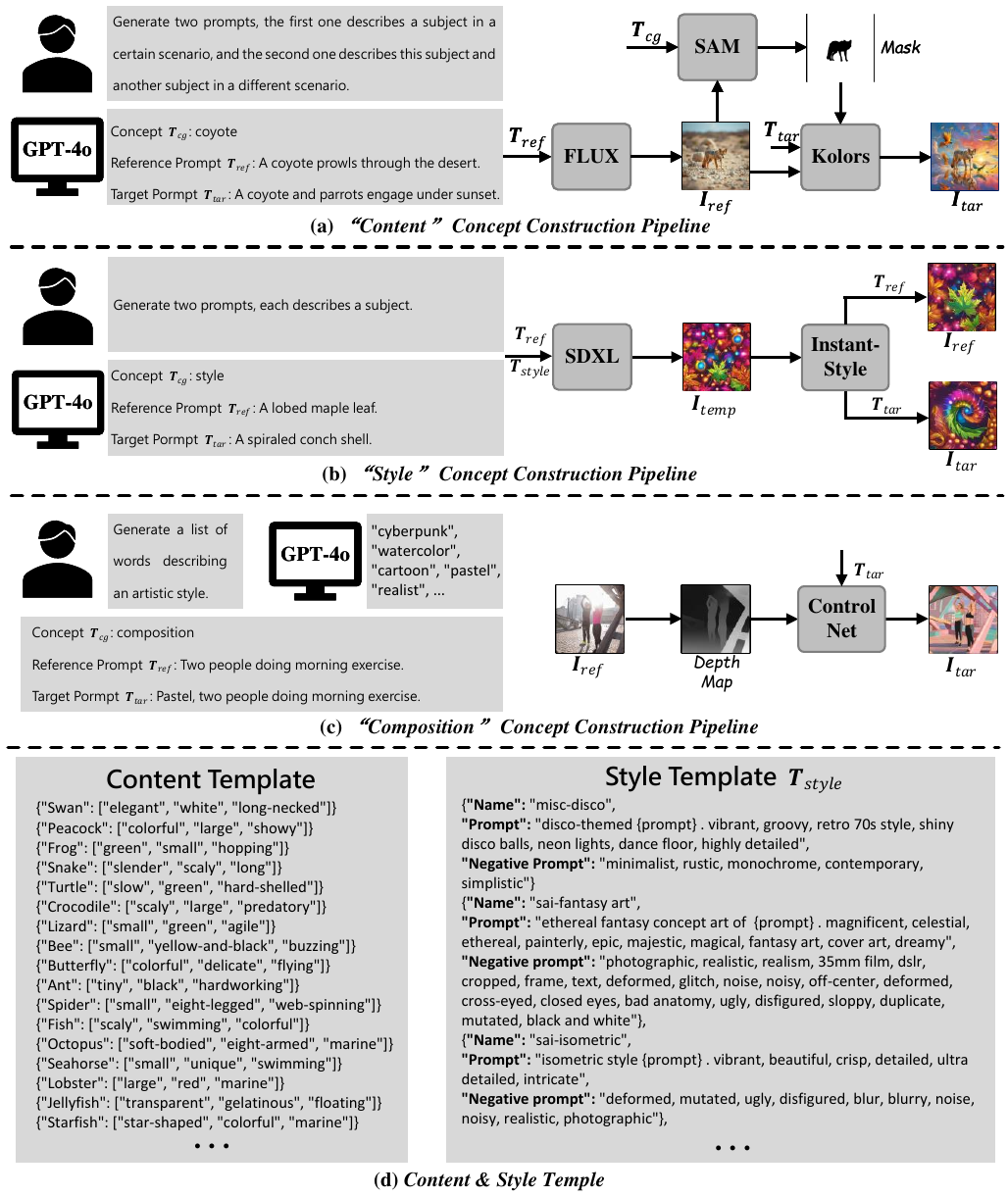}
    \caption{%
        \textbf{Construction Pipeline of our PCD-200K. }We design three data construction pipelines for the three concepts of ``content", ``style", and ``composition", each pipeline uses GPT-4o to obtain reference prompts $\bm{T}_{ref}$, target prompts $\bm{T}_{tar}$, and concept guidance $\bm{T}_{cg}$, and use different models to generate corresponding reference images $\bm{I}_{ref}$ and target images $\bm{I}_{tar}$.
    }
    \label{fig:data_pipe}
    \vspace{5mm}
\end{figure*}

The three visual concepts (subject, style and layout) in our paper are orthogonal and complete for an image as basic concepts, other concepts are belong to them, \textit{e.g.}, material, color to style, or relation, pose to layout.
We can extract fine-grained concepts through our data-driven method. Here we take material-style as example and create a Material-20K dataset and tune it as our paper setting. To get Material-20K, we follow In-Context-LORA \cite{lhhuang2024iclora} and use FLUX \footnote{https://github.com/black-forest-labs/flux} to generate a pair image based on the specified material words. The image prompt and material words are generated by GPT-4~\cite{achiam2023gpt}.
As shown in \cref{fig:met}, the origin version can generate the same material using `style' as guidance. After tuning on Material-20K, we can generate more refined material with `material' as concept guidance.

\vspace{-10pt}

\section{Paired Concept Disentangled Datasets}
\label{sec:sup_dataset}
We disentangle the concepts in an image into three components: content, style, and layout. For each of these concepts, we design distinct data construction pipeline, as shown in the \cref{fig:data_pipe}. Each pipeline first generates prompts describing the same concept and then generates the corresponding image. All images are produced at a resolution of 1024 $\times$ 1024.The datasets for content, style, and layout contain 120K, 40K, and 40K samples, respectively.

\subsection{Content Set}
\noindent \textbf{Prompt Generation. }For the content concepts, our primary objective is to disentangle the various semantic subjects in the image, which can be described by class names. To ensure data diversity, we first use GPT-4 to generate a content template with 120 semantic subjects, \textit{e.g.} humans, animals, plants, furniture, buildings. Three adjectives are generated for each semantic subject to enhance diversity. We then form the concept guidance $\bm{T}_{cg}$ by randomly selecting a subject and an adjective and query GPT-4 to generate two prompts. The reference prompt $\bm{T}_{ref}$ describes the subject in a certain scenario, and the target prompt $\bm{T}_{tar}$ describes the subject and an additional subject in different scenarios.

\noindent \textbf{Image Generation. }We use FLUX to generate reference image $\bm{I}_{ref}$ using $\bm{T}_{ref}$, and use Segment Anything Model (SAM) \cite{kirillov2023segany} to generate mask of the desired concept in $\bm{I}_{ref}$ using $\bm{T}_{cg}$. Subsequently, we use Kolors-Inpainting \cite{kolors} to generate target images $\bm{I}_{tar}$ based on the target prompt $\bm{T}_{tar}$.

\subsection{Style Set}
\noindent \textbf{Prompt Generation. }Describing the style concept in an image with precise language is difficult. Therefore, we uniformly set the concept guidance $\bm{T}_{cg}$ for all samples to ``style" and use GPT-4o to generate two prompts $\bm{T}_{ref}$ and $\bm{T}_{tar}$ describing a random subject in a certain scene. Before generating the image, we randomly select a style description $\bm{T}_{style}$ from the style template which includes 106 common styles in \textit{SDXL Prompt Styler} \footnote{https://github.com/twri/sdxl\_prompt\_styler} and add it to $\bm{T}_{ref}$ as a style guidance.

\noindent \textbf{Image Generation. }Existing stylization methods often deviate from the reference style, and images generated by the same method tend to be similar. Therefore, we first use SDXL \cite{podell2023sdxl} to generate a template image $\bm{I}_{tmp}$ based on $\bm{T}_{ref}$ and $\bm{T}_{style}$, and then use Instant-Style \cite{wang2024instantstyle} to generate $\bm{I}_{ref}$ and  $\bm{I}_{tar}$ based on $\bm{T}_{ref}$ and $\bm{T}_{tar}$, to ensure that the two images have similar styles. The concept guidance for all samples is set to ``layout".

\subsection{Layout Set}
\noindent \textbf{Prompt Generation. }Layout in an image refers to concepts such as relationships, camera angles, or human poses. We collect 50,000 images with a resolution of more than 768 $\times$ 768 and corresponding prompts from Laion Aesthetics 6.5+ as $\bm{I}_{ref}$ and $\bm{T}_{ref}$. Since prompts generated by the language model may conflict with the original layout, we use GPT-4o to generate 100 simple style descriptors, which are added to $\bm{T}_{ref}$ to form $\bm{T}_{tar}$.

\noindent \textbf{Image Generation. }To generate the target image $\bm{I}_{tar}$, we first estimate the depth map of $\bm{I}_{ref}$, and then use ControlNet-Depth \cite{zhang2023adding} to generate $\bm{I}_{tar}$ based on $\bm{T}_{tar}$.

\section{Experimental Details}
\label{sec:sup_exp}

\subsection{Metric Details}
\noindent \textbf{CLIP-I \& CLIP-T. } CLIP \cite{radford2021learning} is a multimodal contrastivse learning model featuring an aligned image-language representation space. Following previous works, we measure the subject consistency by calculating the cosine similarity between the generated image and the reference image in the CLIP space, termed CLIP-I. To assess the fidelity of the text, we computed the cosine similarity between the generated image and the prompt in the CLIP space, termed CLIP - T. We adopt the CLIP ViT-B-32 in all evaluations.

\noindent \textbf{DINO. }DINO \cite{zhang2023dino} is a self-supervised learning method which learns efficient image representations from unlabeled data. Following previous works, we measure the subject consistency by calculating the cosine similarity between the generated image and the reference image in the DINO space. We utilize the DINO-V2-giant in all evaluations.

\noindent \textbf{Mask CLIP-I \& Mask DINO. }To measure the consistency of the disentangled content concept, we mask the specified content concept in the reference image. Then, we calculate the cosine similarities between the generated image and the masked reference image in the CLIP and DINO spaces respectively.

\noindent \textbf{Style Similarity. }To measure the consistency of the disentangled style concept, we adopt the CSD model proposed by Somepalli et al. \cite{somepalli2024measuring}, which learns high-performance style representations through both self-supervised and supervised objectives. We calculate the cosine similarity between the reference image and the generated image in the CSD space.

\noindent \textbf{SSD. }Since the layout concept in image is difficult to measure in the pixel level, we adopt the salient structure distance \cite{park2012modeling} (SSD), which calculate MSE of the saliency maps of reference image and generated image to measure the consistency of layout. 

\noindent \textbf{Aesthetic Score. } We use Aesthetic Score~\cite{christoph2022laion} to evaluate the visual quality of generated images using LAION-Aesthetics Predictor V2\footnote{https://github.com/christophschuhmann/improved-aesthetic-predictor}.

\subsection{Comparative Methods}
Here we give the details of our comparative method:

\noindent \textbf{IP-Adapter} \cite{ye2023ip} injects images into a set of additional cross-attention layers in U-Net to learn image prompts without affecting the original model's capabilities. We utilize their official implementation and set the image scale to 0.5 to achieve a balance between image and text prompts. For a fair comparison, we use their SDXL-Plus version. All concept extraction are guided only by text prompt.

\noindent \textbf{BLIP-Diffusion} \cite{li2023blip} uses BLIP2 \cite{li2024blip} to integrate image and text features for subject customization. We use their official implementation in \textit{Diffusers}\footnote{https://github.com/huggingface/diffusers} based on stable diffusion (SD) 1.5 according to their latest version. The concept guidance is the same as that used in our method.

\noindent \textbf{DEADiff} \cite{qi2024deadiff} uses BLIP2 to learn content and style features in an image, injecting the style features into the style-specific blocks in U-Net to achieve stylization. We use their official implementation based on SD 1.5. For content and style extraction, the guidance are "content" and "style" respectively, and layout extraction uses "content" due to the absence of corresponding settings.

\noindent \textbf{SSR-Encoder} \cite{zhang2024ssr} uses cross-attention to align image features with text prompts to achieve subject extraction. We utilize their implementation based on SD 1.5. The concept guidance is the same as that used in our method.

\noindent \textbf{MS-Diffusion} \cite{wang2025msdiffusion} uses grounding resampler and multi-subject cross-attention mechanisms for smoothly blending multiple subjects into a single image. We utilize their implementation based on SDXL. The input boxes are set to [0.25, 0.25, 0.75, 0.75] for content and layout concept generation, and [0.1, 0.1, 0.9, 0.9] for style concept gengeration.

\noindent \textbf{OmniGen} \cite{xiao2024omnigen} designs a unified image generation model for diverse conditional image generation based on multimodal diffusion transformer (DIT). We utilize their implementation, and add \textit{``the $\langle$concept$\rangle$ is the $\langle$concept$\rangle$ in $\langle$img$\rangle$$\langle|$image\_1$|\rangle$$\langle$/img$\rangle$''} after prompts as their setting.

\section{More Experiments}
\label{sec:sup_analy}

\subsection{Compatibility to Base Models}
\begin{figure}[t]
    \centering
    %\vspace{-10mm}
    \includegraphics[width=0.47\textwidth]{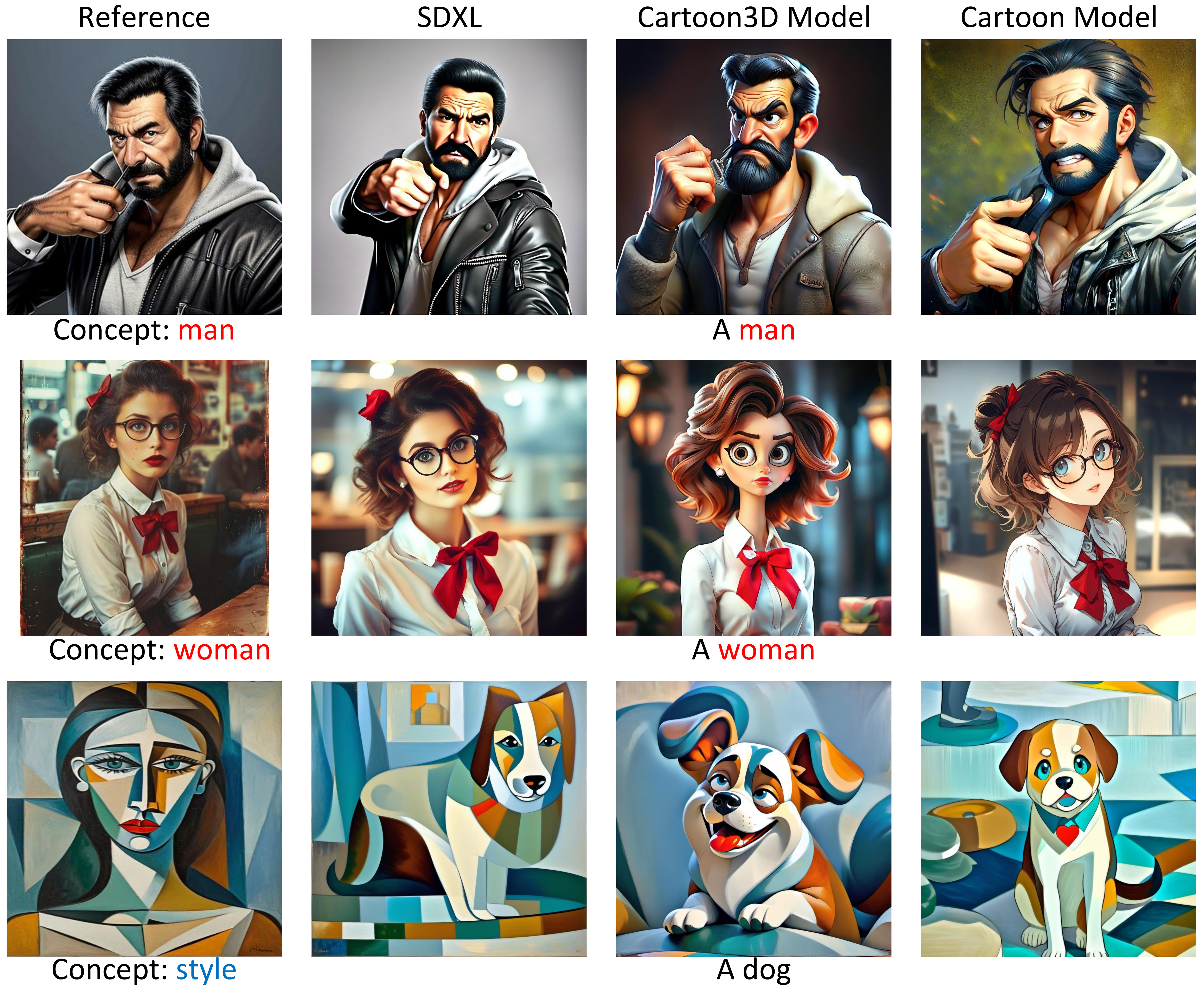}
    %\vspace{-5pt}
    \caption{%
    \textbf{Compatibility to base model.} We change the base model from SDXL to a 3D cartoon model and a cartoon model of the same architecture, demonstrating the compatibility of our \textbf{OmniPrism} to different base models.
    }
    \label{fig:diff_model}
    % \vspace{-10pt}
\end{figure}

\textbf{OmniPrism} is trained using SDXL-1.0 \cite{podell2023sdxl} as base model. During inference, the base model can be flexibly replaced with other models of the same architecture to achieve different effects. We replace SDXL with \textit{Samaritan 3d Cartoon}\footnote{https://civitai.com/models/81270/samaritan-3d-cartoon?modelVersionId=144566} and \textit{HimawariMix}\footnote{https://civitai.com/models/131611/himawarimix?modelVersionId=558064}, which can generate 3d cartoon or cartoon results respectively. As shown in \cref{fig:diff_model}, our \textbf{OmniPrism} demonstrates effective concept disentanglement generation capabilities across different base models. This proves that our method is independent of the base model and can be further transferred to more advanced models, such as FLUX.

\subsection{Additional Comparison with Other Works}
\begin{figure}[t]
    \centering
    %\vspace{-10mm}
    \includegraphics[width=0.47\textwidth]{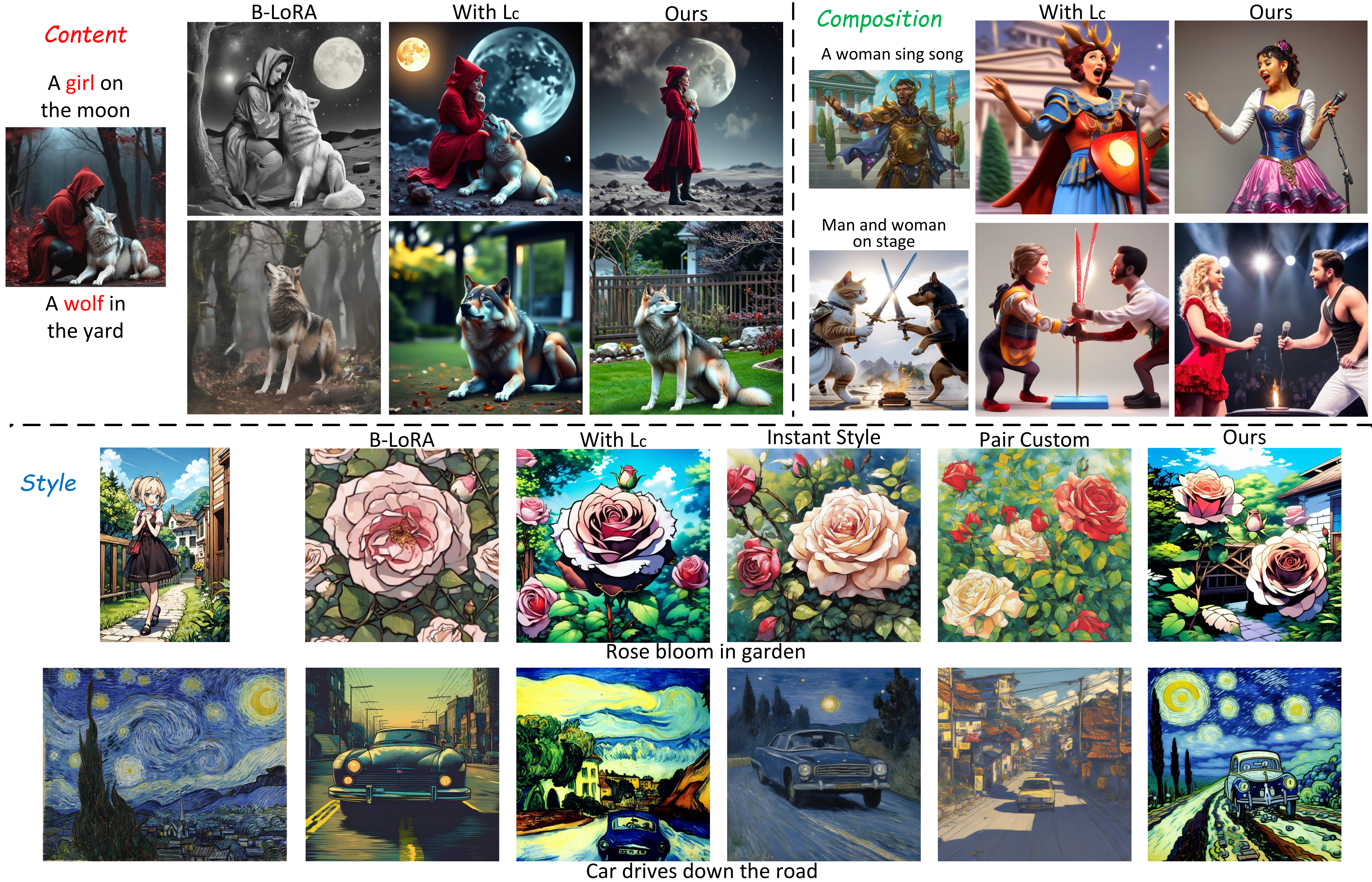}
    %\vspace{-5pt}
    \caption{%
    \textbf{Additional Comparison with Other Methods.}
    }
    \label{fig:obj}
    % \vspace{-10pt}
\end{figure}

In the main paper, we compare some works that is highly related to ours. Here we add some additional comparisons with other works, including B-LoRA~\cite{frenkel2024implicit} for content and style concept, Instant Style~\cite{wang2024instantstyle} and Pair Custom~\cite{jones2024customizing} for style concept, and Yao et~al.~\cite{yao2023multi} for all concepts. Yao et~al. design a multi-view disentanglement approach to learns disentangled representations. To be applicable for generation, we refer to their theory and change our $\mathcal{L}_{COD}$ to:
\begin{equation}\label{eq:class_free_exp}
\mathcal{L}_{C} = \rvert\rvert\bm{f}_{cpt} - \bm{f}^{tar}_{cpt}\rvert\rvert_2 - \bm{H}(\bm{f}_{cpt}),
\end{equation}
as the comparison. As shown in~\cref{fig:obj} and~\cref{table:sup_quan}, our method achieve better content and style consistency. B-LoRA needs extra training for each sample, which limits their application. Instant Style can only disentangle style concepts. Pair Custom learns the difference between content-style pairs to achieve stylization, which is limited in common scenarios with only style images. While we learn text-guided concept disentanglement, which not only adapt to diverse scenarios, but also achieves better concept consistency and text fidelity. Yao et~al. do not impose constraints on the mutual independence of different concepts, and thus concept confusion is prone to occur when applied to generation. In contrast, our method solves these problems and achieve better concept disentangling performance.

\begin{table}[t]
\centering
\caption{Additional Comparison with Other Methods.}
\scalebox{0.85}{
\begin{tabular}{@{}lcccccc@{}}
\toprule
Method          & \makecell[c]{Mask\\CLIP-I $\uparrow$} & \makecell[c]{Mask\\DINO $\uparrow$}  & CLIP-T $\uparrow$  & \makecell[c]{Style\\Similarity $\uparrow$} & SSD $\downarrow$  & \makecell[c]{Aesthetic\\Score $\uparrow$}       \\ \midrule
B-LoRA       & 0.695      & 0.236     & 0.262            & 0.360 & --   & 6.433              \\
Instant-Style   & --   & --      & 0.278             & 0.548     & --   & \textbf{6.650}                  \\
Pair-Custom   & --    & --    & \textbf{0.303}              & 0.355    & --     & 6.425                \\
with $\mathcal{L}_C$         & 0.785   &  \underline{0.441}     & 0.280             & 0.508   & 0.287     & 6.233                  \\
Ours            & \textbf{0.797} & \textbf{0.513} & \underline{0.296}     & \textbf{0.585}    & \textbf{0.267}    & \underline{6.485}      \\ \bottomrule
\end{tabular}
}

\label{table:sup_quan}
% \vspace{-5mm}
\end{table}

\subsection{Evaluation on Disentangling Capability}
All our evaluations in the main paper are conducted on generated images. To further evaluate the capability to disentangle intermediate concept representations, we refer to Yao et~al.~\cite{yao2023multi} and calculate the DCI disentanglement scores~\cite{eastwood2018framework} on MPI3D complex~\cite{gondal2019transfer}. Similar to Section 3.C in the main paper, we do not evaluate OmniGen because it lacks learnable concept representations. We perform PCA prior to regression. As shown in~\cref{table:sup_dci}, our model achieves the best concept disentanglement.

\begin{table}[ht]
% \vspace{-3mm}
\caption{Evaluation on DCI disentanglement scores.}
\scalebox{0.6}{
\begin{tabular}{@{}lccccccc@{}}
\toprule
Method          & IP-Adapter & BLIP-Diffusion  & DEADiff  & SSR-Encoder & MS-Diffusion & Yao et~al. & Ours       \\ \midrule
DCI  $\uparrow$    & 0.422$\pm$0.003     & 0.358$\pm$0.002     & 0.336$\pm$0.002   & \underline{0.429$\pm$0.002} & 0.422$\pm$0.002 & 0.420$\pm$0.020  & \textbf{0.507$\pm$0.004}            \\\bottomrule
\end{tabular}
}
\label{table:sup_dci}
% \vspace{-5mm}
\end{table}

\subsection{Two Content Combination}
\begin{figure}[t]
    \centering
    \includegraphics[width=0.47\textwidth]{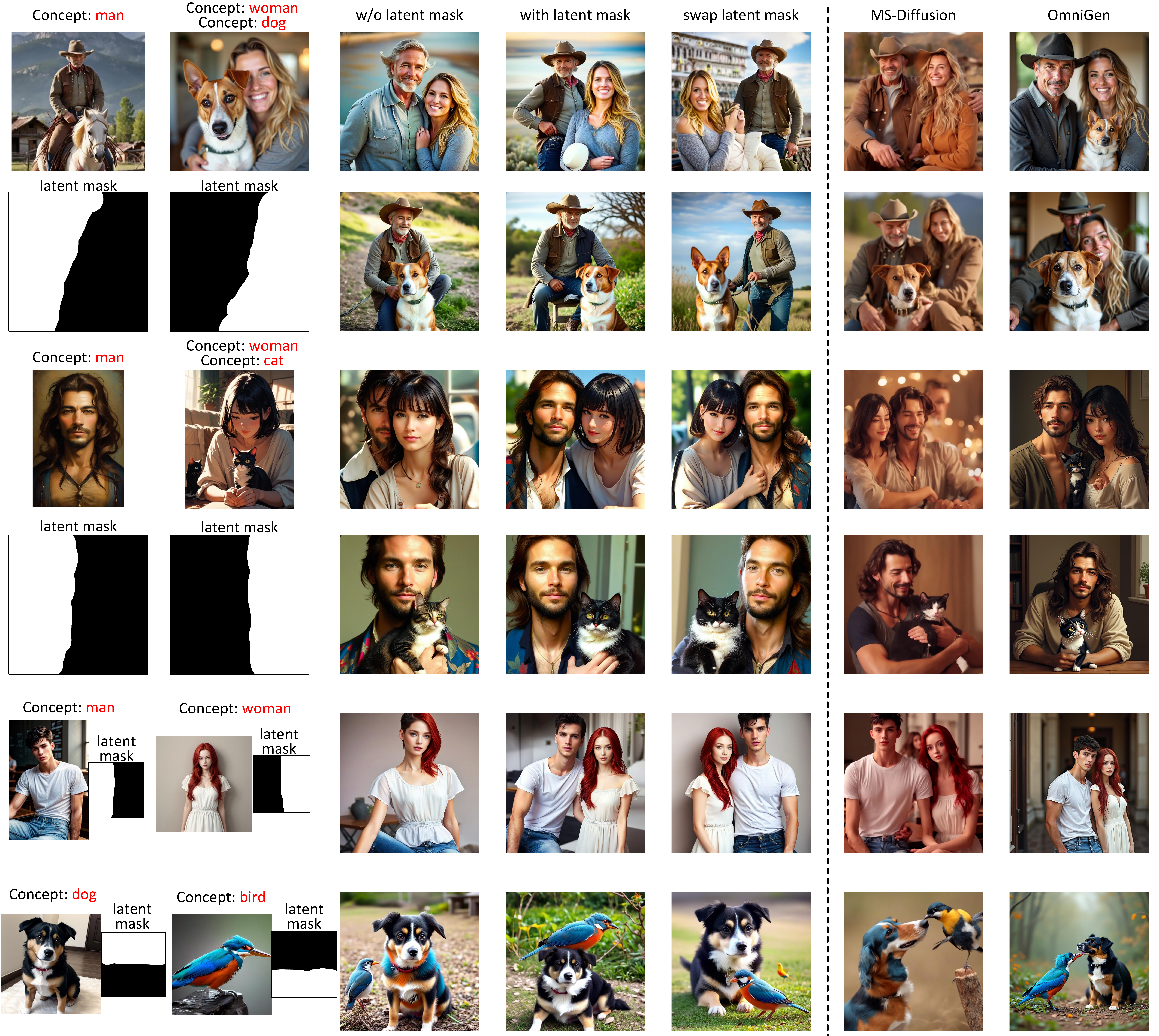}
    \caption{%
        \textbf{Combination of two content concepts.} We use latent masks to assign layouts to different concepts to prevent conflicting.
    }
    \label{fig:multi_obj}
    % \vspace{-5pt}
\end{figure}

In our paper, we demonstrate the creative generation results achieved by combining various concepts, such as content and style, to achieve subject stylization. The same concept, such as multiple style or composition concepts, is difficult to combine due to they may conflict with each other. Combining multiple content concepts together is also prone to concept mix-up, as the model may struggle to differentiate between subjects. To address this, we refer to IP-Adapter and add masks in the latent space to assign different layouts to each subject, as shown in \cref{fig:multi_obj}. We also show the results of MS-Diffusion and OmniGen in combining two content. They are sometimes disturbed by irrelevant concepts, resulting in concept leakage.

\subsection{Multi-Content Disentanglement and Combination}
\begin{figure}[t]
    \centering
    \includegraphics[width=0.47\textwidth]{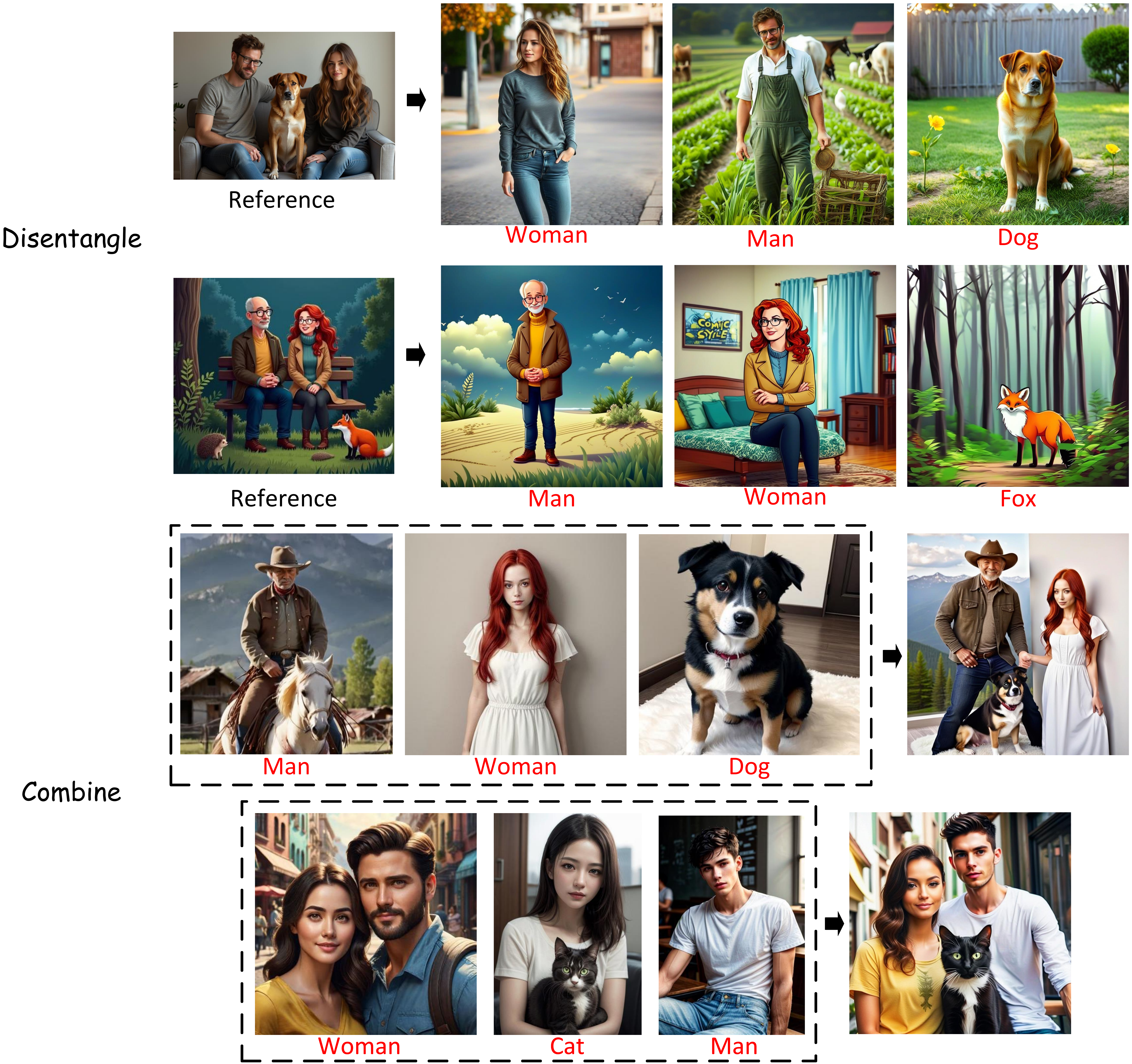}
    \caption{%
        \textbf{Disentanglement and combination of multiple content concepts.}
    }
    \label{fig:3obj}
    % \vspace{-5pt}
\end{figure}

As the number of subjects to be disentangled or combined increases, the difficulty of generating high-similarity images also rises. Here, we respectively present the results of disentangling and generating individual subject concepts from scenarios containing multiple objects, as well as combining multiple subjects into a single image, as shown in~\cref{fig:3obj}. It can be observed that our model exhibits excellent disentangling and combining capabilities even in scenarios involving multiple objects.

\subsection{Complex Prompts for Generation}
\begin{figure}[t]
    \centering
    \includegraphics[width=0.47\textwidth]{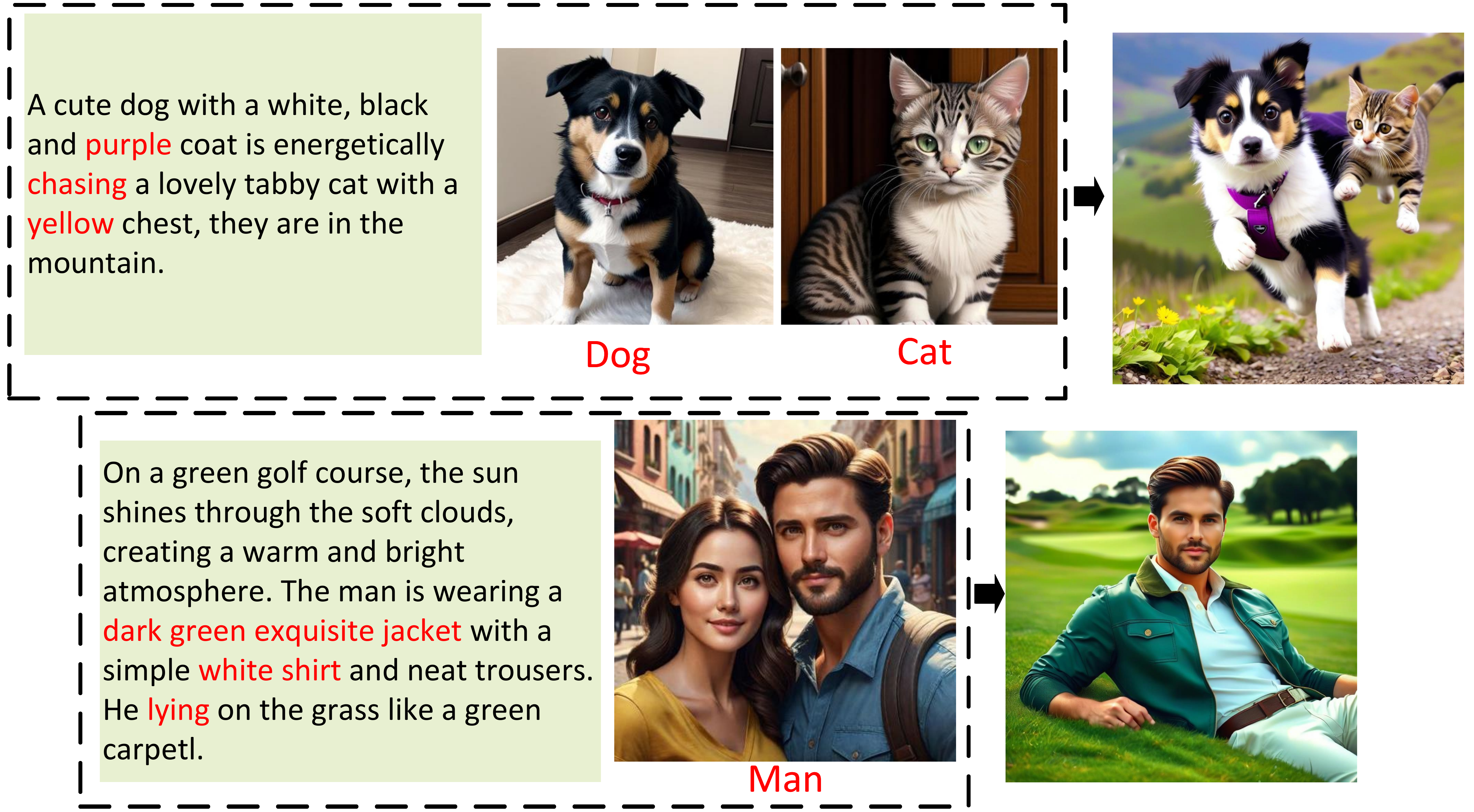}
    \caption{%
        \textbf{Disentanglement and combination of multiple content concepts.}
    }
    \label{fig:complex}
    % \vspace{-5pt}
\end{figure}

When using complex prompts to control visual concept generation, a core challenge lies in balancing the two to achieve high text fidelity and high conceptual similarity. Here, we present the generation results of some complex prompts. As shown in~\cref{fig:complex}, we marked the elements we want in the prompt in red. The generated result can well preserve the concept of the reference image and the elements mentioned in the prompt, yielding high-quality and balanced generation.

\section{More Qualitative Results}
\label{sec:sup_res}
\begin{figure*}[t]
    \centering
    \includegraphics[width=0.95\textwidth]{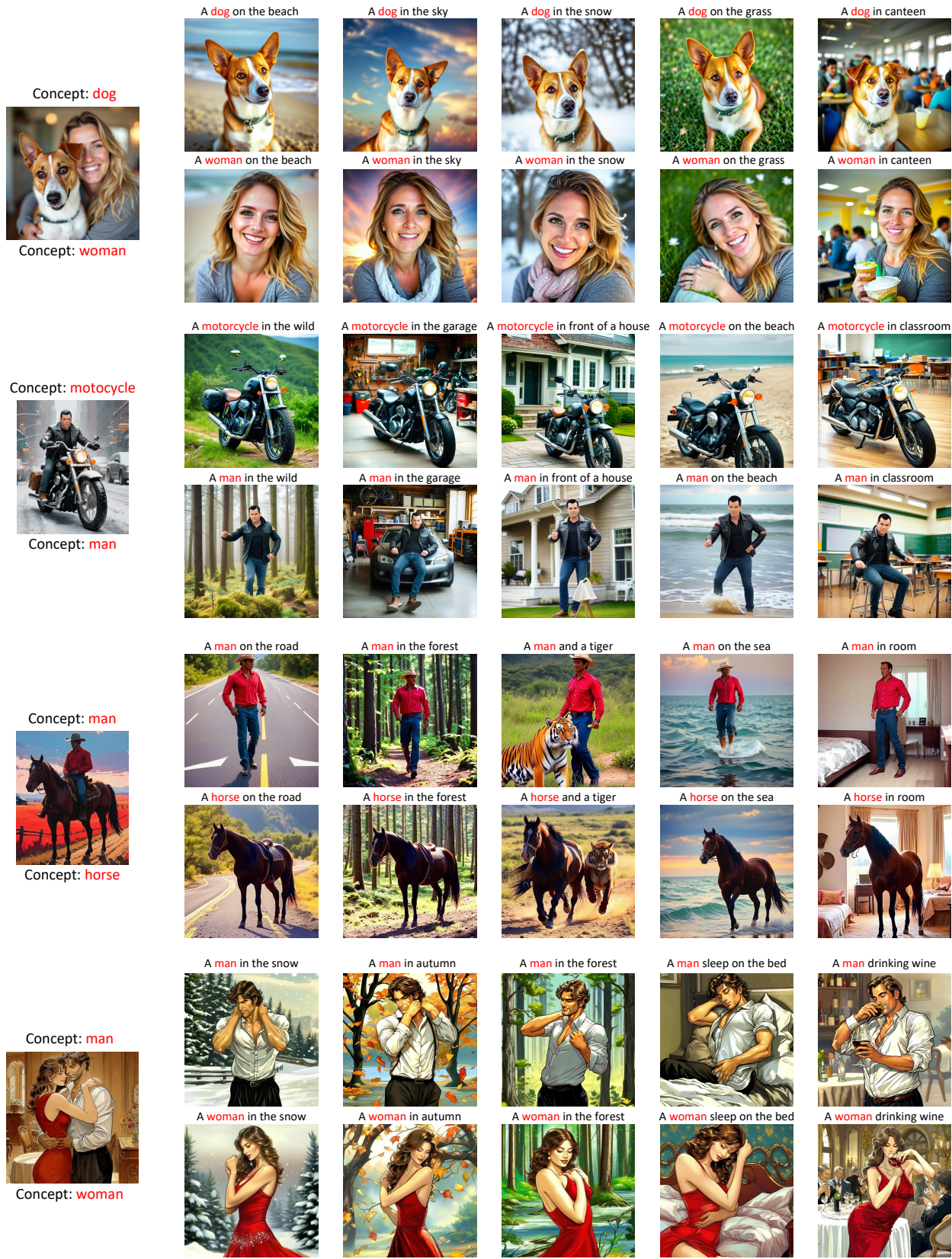}
    \caption{%
        \textbf{Disentangled Generation of Content.}
    }
    \label{fig:sup_obj}
\end{figure*}

\begin{figure*}[t]
    \centering
    \includegraphics[width=1.0\textwidth]{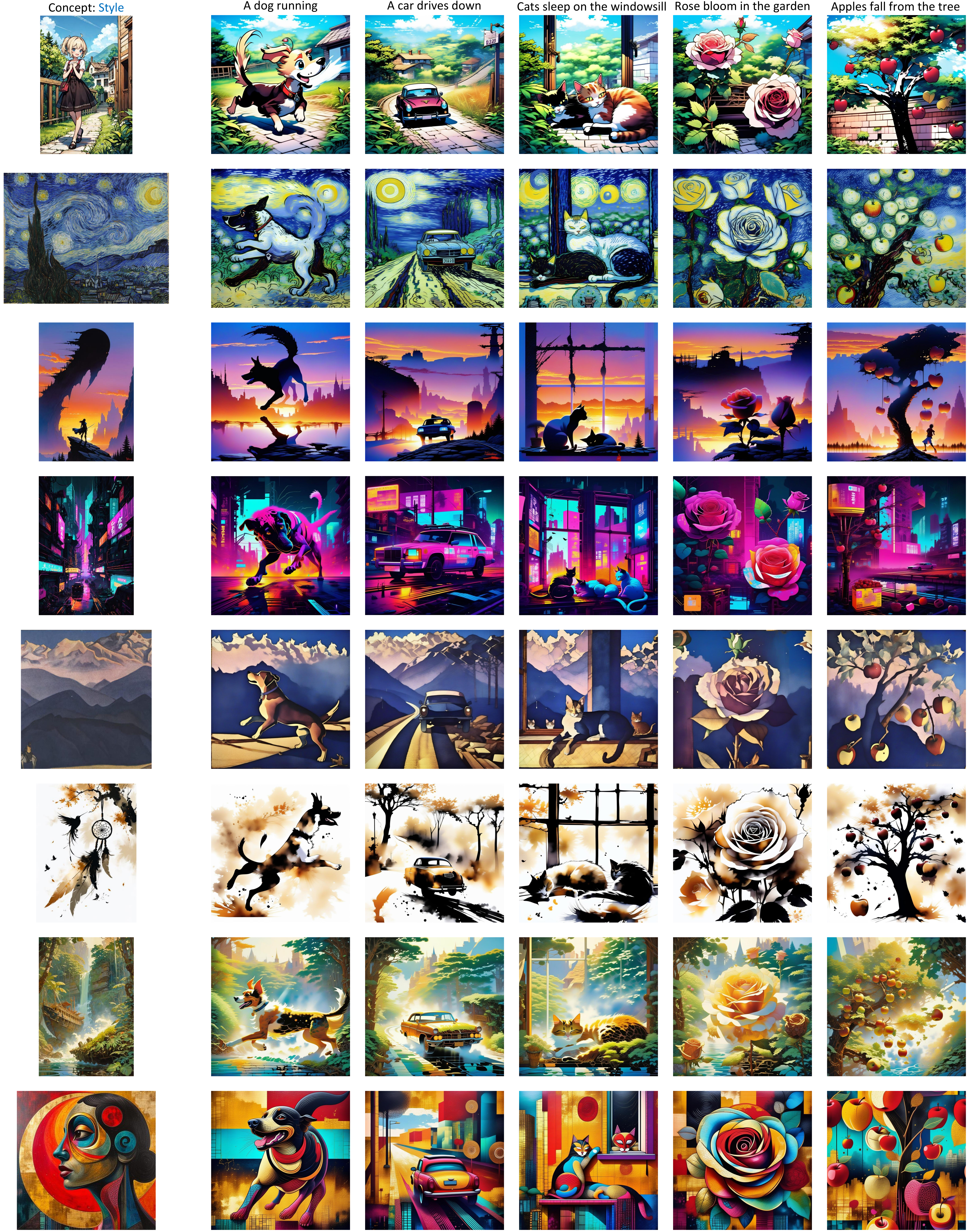}
    \caption{%
        \textbf{Disentangled Generation of Style.}
    }
    \label{fig:sup_style}
\end{figure*}

\begin{figure*}[ht]
    \centering
    \includegraphics[width=1.0\textwidth]{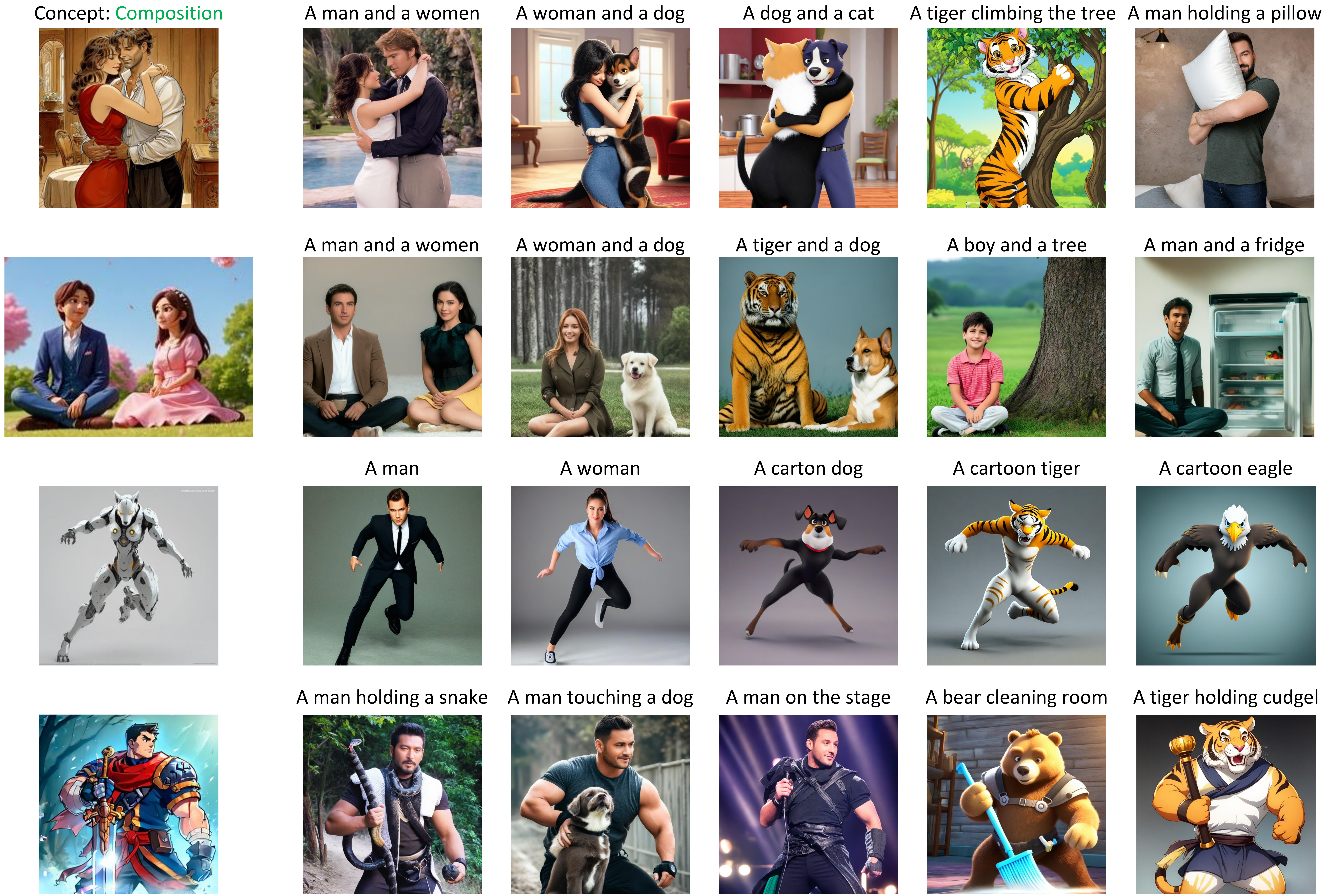}
    \caption{%
        \textbf{Disentangled Generation of Composition.}
    }
    \label{fig:sup_com}
\end{figure*}

We present the generation results for various concepts on more reference images, ``content" concepts are shown in \cref{fig:sup_obj}, ``style" concepts are shown in \cref{fig:sup_style}, and ``composition" concepts are shown in \cref{fig:sup_com}. These results further demonstrate the powerful concept disentangling generation capability of our \textbf{OmniPrism}.

\section{Social Impact}
\label{sec:sup_impact}
The proposed \textbf{OmniPrism} has the potential to revolutionize creative industries by enabling artists and designers to explore new styles and democratize content creation for small businesses and independent creators.
However, it also poses risks such as the creation of realistic but false content that can spread misinformation and deepfakes, potentially undermining public trust and political discourse.
The unauthorized use of copyrighted material raises legal and ethical concerns, while biases in training datasets can perpetuate harmful stereotypes and marginalize certain groups.
Any user-facing application based on this method needs to establish a security review mechanism to filter out potentially harmful content.

\vfill

\end{document}